# DeepFAN, a transformer-based deep learning model for human-artificial intelligence collaborative assessment of incidental pulmonary nodules in CT scans: a multi-reader, multi-case trial

**Zhenchen Zhu[1#], Ge Hu[2#], Weixiong Tan[3#], Kai Gao[3#], Chao Sun[4#], Zhen Zhou[3#], Kepei Xu[1], Wei Han[5], Meixia Shang[6], Xiaoming Qiu[7,8], Yiqing Tan[9], Jinhua Wang[1], Zhoumeng Ying[1,10], Li Peng[1,11], Wei Song[1], Lan Song[1*], Zhengyu Jin[1*], Nan Hong[4*], Yizhou Yu[12*]**

[1] Department of Radiology, State Key Laboratory of Complex Severe and Rare Diseases, Peking Union Medical College Hospital, Chinese Academy of Medical Sciences and Peking Union Medical College, Beijing, China.

[2] Theranostics and Translational Research Center, National Infrastructures for Translational Medicine, Institute of Clinical Medicine, State Key Laboratory of Complex Severe and Rare Diseases, Peking Union Medical College Hospital, Chinese Academy of Medical Sciences and Peking Union Medical College, Beijing, China.

[3] Artificial Intelligence Lab, Deepwise Healthcare, Beijing, China.

[4] Department of Radiology, Peking University People's Hospital, Beijing, China.

[5] Department of Epidemiology and Health Statistics, Institute of Basic Medicine Sciences, Chinese Academy of Medical Sciences & Peking Union Medical College, Beijing, China.

[6] Department of Biostatistics, Peking University First Hospital, Beijing, China.

[7] Department of Radiology, Huangshi Central Hospital, Affiliated Hospital of Hubei Polytechnic University, Hubei Province, China.

[8] Key Laboratory of Cerebrovascular Disease Imaging and Artificial Intelligence, Huangshi，Hubei Province, China

[9] Department of Radiology, Wuhan Third Hospital, Tongren Hospital of Wuhan University, Wuhan, Hubei Province, China.

[10] 4+4 Medical Doctor Program, Chinese Academy of Medical Sciences & Peking Union Medical College, Beijing, China.

[11] Department of Medicine Imaging, School of Clinical Medicine, Southwest Medical University, Luzhou, China.

[12] School of Computing and Data Science, The University of Hong Kong, Hong Kong SAR, China.

#These authors contributed equally: Zhenchen Zhu, Ge Hu, Weixiong Tan, Kai Gao, Chao Sun and Zhen Zhou.
*Corresponding authors: correspondence to Lan Song, Zhengyu Jin, Nan Hong and Yizhou Yu.

## Abstract

The widespread adoption of CT has notably increased the number of detected lung nodules. However, current deep learning methods for classifying benign and malignant nodules often fail to comprehensively integrate global and local features, and most of them have not been validated through clinical trials. To address this, we developed DeepFAN, a transformer-based model trained on over 10K pathology-confirmed nodules and further conducted a multi-reader, multi-case clinical trial to evaluate its efficacy in assisting junior radiologists. DeepFAN achieved diagnostic area under the curve (AUC) of 0.939 (95% CI 0.930-0.948) on an internal test set and 0.954 (95% CI 0.934-0.973) on the clinical trial dataset involving 400 cases across three independent medical institutions. Explainability analysis indicated higher contributions from global than local features. Twelve readers' average performance significantly improved by 10.9% (95% CI 8.3%-13.5%) in AUC, 10.0% (95% CI 8.9%-11.1%) in accuracy, 7.6% (95% CI 6.1%-9.2%) in sensitivity, and 12.6% (95% CI 10.9%-14.3%) in specificity (*P*<0.001 for all). Nodule-level inter-reader diagnostic consistency improved from fair to moderate (overall $\kappa$: 0.313 vs. 0.421; *P*=0.019). In conclusion, DeepFAN effectively assisted junior radiologists and may help homogenize diagnostic quality and reduce unnecessary follow-up of indeterminate pulmonary nodules. Chinese Clinical Trial Registry: ChiCTR2400084624.

**Introduction**

The 2022 global cancer statistics reveal that lung cancer remains the leading cause of cancer-related deaths worldwide, with an estimated 1.8 million deaths annually[1]. In China, lung cancer is not only the most common (over 1.0 million new cases annually) but also the most fatal cancer, accounting for 28.5% of all cancer-related deaths, far above the global average rate (18.7%)[1,2]. The accumulated economic burden associated with lung cancer was estimated to be 25,069 million USD in China in 2017 (0.121% of Gross Domestic Product) and was expected to increase over the years[3]. This economic burden is further exacerbated for patients in less developed regions due to socio-economic disparities, leading to delayed diagnoses and poorer prognoses[4]. Therefore, timely identification and early diagnosis of lung cancer are crucial for mitigating this burden.

Computed tomography (CT) is the main imaging technique for identifying lung cancer at an early stage. As chest CT becomes more affordable and available, the increased use of chest CT led to the detection of millions of incidental pulmonary nodules (IPNs) annually, and this number further increases with the implementation of lung cancer screening[5-8]. The vast number of CT images has tremendously increased radiologists' workload, often forcing them to reduce the time spent on each case, thereby leading to an increase in interpretive errors[9]. Although the majority of nodules are ultimately classified as false positive findings for lung cancer[10], accurately determining whether a small nodule is malignant at the time of initial detection remains a significant challenge. Additionally, given that image features on CT are often subject to inter-observer inconsistency, which is highly affected by radiologists' working experience, the assessment of IPNs in the real world might be imprecise and inconsistent, resulting in heightened patient anxiety that complicates the clinical process for managing IPNs[11-13].

Therefore, an accurate method/tool to help radiologists efficiently and consistently assess pulmonary nodules is urgently needed. The Mayo model and Brock model are well-known risk models for evaluating the malignancy of solitary pulmonary nodules, but they were developed using Western populations and have shown only moderate performance in the context of IPNs[14,15]. Additionally, patient demographic information and smoking history might be unavailable for IPNs in real-world settings. Recently, artificial intelligence (AI)-assisted diagnostic tools have shown excellent performance in classifying IPNs as malignant or benign, reaching a level comparable to skilled radiologists[16-22]. However, these models were mainly developed on lung cancer screening populations with predominantly low-dose chest CT scans, which did not fully represent non-screening CT scans obtained during physical examinations and other clinical purposes[23], and most of them were not verified in the clinical practices. Besides, current AI models for pulmonary nodule assessment predominantly rely on convolutional neural networks (CNNs) and their derivatives, inherently emphasizing local feature extraction owing to the convolution operation[18,19,22,24]. To address the above challenges, we propose a deep feature aggregation network (DeepFAN) and test this method in a rigorous clinical setting. This novel framework is built on Vision Transformers (ViT)[25]—a self-attention based deep neural architecture for computer vision tasks, to capture global features of nodules by leveraging global attention passes across the entire input image. Additionally, a fine-grained CNN model is integrated into our framework to extract detailed local features of nodules. Finally, a graph convolution network (GCN)[26] is adopted to effectively amalgamate global and local features. This approach facilitates relation learning and surpasses traditional feature fusion methods[26].

While the performance metrics of AI-assisted tools as standalone devices are important, their value also lies in how much they enhance radiologists' performance in a human-AI collaborative mode in daily clinical practice. The multireader multicase (MRMC) study design is presently the most recognized method for analyzing the performance of human readers assisted by AI-based solutions, but MRMC studies have a time-consuming and costly nature[27-29]. Kim and his colleagues conducted a simplified MRMC study without a washout period to demonstrate the efficacy of an AI-based computer-aided diagnosis tool in improving both diagnostic performance of indeterminate pulmonary nodules detected in chest CT scans and interobserver agreement for risk stratification[30]. However, a rigorous MRMC study based clinical trial on Asian populations from developing countries for an AI-assisted tool for the diagnosis of pulmonary nodules detected in chest CT scans is still absent. Meanwhile, there is significant interest in factors (e.g. accuracy of AI system, susceptibility to AI, personality of radiologists, and working experience) that can influence radiologists' diagnostic decisions during AI-assisted reading sessions in order to facilitate the incorporation of AI-assisted tools into clinical practice[31-33]. Therefore, this study collected a large cohort of IPN patients, more than 10K pathological confirmed pulmonary nodules from nine medical institutions in China, to train the proposed DeepFAN model to differentiate malignant pulmonary nodules from benign ones, and then conduct the first rigorous clinical MRMC-study based national clinical trial (ChiCTR2400084624) in another three independent medical institutions in China with diverse population sizes (Peking University People's Hospital, 2022PHA118-001; Wuhan Third Hospital, Wuhan No.3 QX2023-002; and Huangshi Central Hospital, Lun Kuai Shen [2023] No. 2). Besides, the generalization ability of DeepFAN was tested on the national lung screening trial (NLST) dataset as well as the dataset used in the clinical trial. Furthermore, to better understand the human-AI collaboration process, we explored the influence of various factors on the accuracy of human-AI diagnostic outcomes. Finally, we aimed to provide explanations to divergence within human-AI decision-making by visualizing the analytic process of AI individually as well as human-AI collaboratively.

## Results

### Deep feature aggregation network (DeepFAN)

The Deep Feature Aggregation Network (DeepFAN, version 1.0, owned by Beijing Deepwise & League of PHD Technology Co., Ltd.), the AI model tested in the clinical trial, has been approved by the National Medical Products Administration (NMPA) of China (National Medical Device Approval No. 20243211932). Its neural architecture is detailed in **Figure 1**. DeepFAN integrates three component neural networks that enhance hybrid feature learning and characterization for CT images. The first component is a vision transformer (ViT) that effectively extracts global features of pulmonary nodules and their surrounding areas, encoding the overall morphology and context of a nodule. The second component is a three-dimensional residual network with counterfactual attention learning and an attention dropout layer (CAL-ADL 3D ResNet)[34,35], which extracts local features representing detailed radiologic characteristics of a nodule (such as density, spiculation, and lobulation). Compared to general convolutional neural networks (CNN), this network excels in fine-grained machine learning by capturing small-scale characteristics, thereby enhancing the AI model's capability in sensing and differentiating subtle nodule features. In the last component, a graph convolutional network (GCN) is adopted to aggregate and fuse the global features of a nodule extracted using the ViT and the local features extracted using the CAL-ADL 3D ResNet, aiming to comprehensively understand pulmonary nodule characterization.

DeepFAN was developed and further evaluated using a dataset of 11,438 pathologically confirmed pulmonary nodules from 8,172 patients, collected from nine hospitals across seven provinces in China, including prominent medical institutions such as Peking Union Medical College Hospital. The dataset was randomly partitioned in a 7:1:2 ratio into training, validation, and internal test sets. The training set consisted of 5,636 patients with 7,873 nodules (1,718 benign and 6,155 malignant), the validation set included 831 patients with 1,216 nodules (254 benign and 962 malignant), and the internal test set comprised 1,705 patients with 2,349 nodules (600 benign and 1,749 malignant). Additionally, to further test its generalization ability, as an extension to the clinical trial, DeepFAN was also validated on the National Lung Screening Trial (NLST) dataset, which includes 7,934 patients with 17,892 nodules (16,821 benign vs. 1,071 malignant). The basic characteristics of patients, nodules and parameters of CT scans and reconstruction in the training, validation, internal test sets and NLST dataset are shown in **Supplementary Table 1-2**.

### Internal validation and generalization ability of DeepFAN

The diagnostic performance of DeepFAN in differentiating malignant nodules from benign ones is shown in **Supplementary Table 3**. On the internal test set, DeepFAN achieved an area under the receiver operating characteristic curve (AUC) of 0.939 (95% confidence interval [CI], 0.930-0.948), along with a sensitivity of 0.953 (0.943-0.962) and a specificity of 0.733 (0.699-0.768).

To assess the contribution of each module within DeepFAN to its overall performance, ablation experiments were conducted on the internal test set using ViT, ResNet50, and CAL-ADL 3D ResNet as the three component networks. The procedure involved progressively removing, adapting, or replacing these components of DeepFAN to evaluate their individual impact on the system efficacy. As illustrated in **Supplementary Table 4**, the combination of ViT, CAL-ADL 3D ResNet, and GCN (model 9=DeepFAN) achieved the highest AUC in ablation experiments, significantly outperforming related baseline methods (model 9 vs. model 1 to 3, $P<0.001$) and other neural architectures (model 9 vs. model 4-8, $P<0.05$).

As a supplement to the clinical trial, DeepFAN was tested on the NLST dataset to further evaluate its generalization ability (**Supplementary Table 3**). Despite the significant reduction in the proportion of malignant nodules in the NLST dataset (with a benign-to-malignant ratio of 15.71 compared to 0.92 in the clinical trial dataset), the AI model's performance under other metrics still met expectations. Specifically, it achieved an AUC of 0.943 (0.933-0.953), sensitivity of 0.889 (0.869-0.908), specificity of 0.897 (0.893-0.902), and accuracy of 0.897 (0.892-0.901). Our DeepFAN exhibited a negative predictive value (NPV) of 0.992 (0.991-0.994) and a positive predictive value (PPV) of 0.356 (0.338-0.374). These results indicate that DeepFAN can maintain excellent generalization performance across different clinical scenarios.

We have also compared the performance of DeepFAN with that of previous methods for pulmonary nodule diagnosis by gathering the performance measures reported in published papers. Except the recently published deep convolutional neural network (DCNN) model[22], the methods presented in the remaining studies—namely, the lung cancer prediction convolutional neural network (LCP-CNN) model, the Mayo model, Brock model, and deep learning (DL) model[14,17,18,36,37]—were rigorously assessed utilizing the NLST dataset, a prominent repository within the realm of lung nodule analysis. **Supplementary Table 5** presents a comparative overview of these approaches. While we endeavored to accurately reproduce the proposed models and dataset handling processes, some discrepancies may remain due to limited available details. Nevertheless, DeepFAN exhibited outstanding performance in assessing IPNs, achieving high AUC, accuracy, and sensitivity. These results suggest that the model could serve as a valuable tool for the preliminary assessment of IPNs, potentially reducing the risk of misdiagnosing malignant nodules.

**Baseline characteristics of clinical trial**

The clinical trial design followed a strict MRMC protocol[27] and the workflow is presented in **Figure 2** and a detailed explanation is illustrated in the Method section. The clinical trial dataset contained 463 pathologically confirmed IPNs (222 benign vs. 241 malignant) from 400 consecutive patients (197 benign vs. 203 malignant) enrolled according to predefined inclusion and exclusion criteria (**Extended Data Figure 1**). Specifically, 166 patients with 204 IPNs were from clinical trial center I, 46 patients with 46 IPNs from center II, and 188 patients with 213 IPNs from center III. Basic characteristics of patients/nodules and parameters of CT scans are shown in **Table 1** and **Supplementary Table 6-7**. The readers participating in the clinical trial are junior radiologists with 1-5 years of working experience, including three readers from clinical trial center I, four from center II, and five from center III. Detailed information about the readers is provided in **Supplementary Table 8**.

**Performance of DeepFAN in clinical trial**

In the clinical trial, DeepFAN alone obtained an average AUC of 0.954 (0.934-0.973), a sensitivity of 0.950 (0.923-0.978), and a specificity of 0.851 (0.805-0.898). More specifically, the AUCs in the three clinical trial centers were 0.947 (0.915-0.978), 0.975 (0.923-1.000), and 0.963 (0.937-0.988), respectively (**Figure 3**). The sensitivities were 0.925 (0.874-0.972), 1.000 (1.000-1.000), and 0.966 (0.932-0.992), while the specificities were 0.878 (0.813-0.938), 0.815 (0.640-0.957), and 0.835 (0.762-0.905), respectively (**Supplementary Table 3**).

**Performance of readers with and without AI assistance**

In the control group of the clinical trial, the readers independently evaluated the malignancy of 463 nodules from the 400 patients (**Supplementary Table 9**). The average AUC of the twelve readers was 0.667 (0.616-0.719),

with a sensitivity of 0.693 (0.676-0.709), a specificity of 0.605 (0.586-0.623), an accuracy of 0.651 (0.638-0.663). The highest AUC, sensitivity, and specificity achieved by the readers were 0.773 (0.731-0.815; reader 12), 0.954 (0.927-0.979), and 0.928 (0.892-0.958; reader 08), while the lowest values were 0.533 (0.480-0.586; reader 06), 0.402 (0.339-0.466; reader 08), and 0.423 (0.364-0.489; reader 09), respectively. At the patient level, the average AUC of the independent reading was 0.733 (0.685-0.780), with a sensitivity of 0.759 (0.743-0.776), a specificity of 0.568 (0.548-0.588), and an accuracy of 0.665 (0.652-0.679).

With the assistance of DeepFAN (**Supplementary Table 9**), the average AUC of the test group was improved to 0.776 (0.733-0.819), sensitivity 0.769 (0.754-0.784), specificity 0.731 (0.714-0.747), accuracy 0.751 (0.739-0.762), PPV 0.756 (0.740-0.772), NPV 0.744 (0.728-0.762), and F1-score 0.762 (0.750-0.775). The highest AUC, sensitivity, and specificity achieved by the AI-assisted readers were 0.883 (0.852-0.914; reader 02), 0.934 (0.903-0.964; reader 12), and 0.959 (0.932-0.985; reader 08), while the lowest values were improved to 0.693 (0.645-0.741; reader 06), 0.531 (0.467-0.591; reader 08), and 0.518 (0.452-0.586; reader 12), respectively. At the patient level, the average AUC of the AI-assisted reading was 0.840 (0.807-0.873), with a sensitivity of 0.833 (0.818-0.848), a specificity of 0.705 (0.687-0.724), and an accuracy of 0.770 (0.758-0.782).

**Figure 3** illustrates the performance of the twelve readers on each clinical trial center dataset. The arrows in the plot highlight the enhancements in reader performance, which are particularly evident in specificity with an average improvement of 0.126 (0.109-0.143). The average improvements in AUC, sensitivity, and accuracy of the twelve readers were 0.109 (0.083–0.135), 0.076 (0.061–0.092), and 0.100 (0.089–0.111), respectively (see Supplementary Table 9). **Extended Data Figure 2** provides further details of the diagnostic indicators for each reader, both with and without AI assistance, and their comparison with DeepFAN performance. DeepFAN outperformed readers with one to five years of experience across most metrics (P < 0.001 for all AUC and accuracy comparisons), and reader performance remained significantly below that of DeepFAN even with AI assistance. A subset of readers, however, achieved sensitivity and specificity comparable to or exceeding those of DeepFAN. This is specifically illustrated in the radar maps (**Figure 4**), where the areas of AI-assisted diagnostic metrics (depicted in blue) are larger than those of independent diagnostic metrics (depicted in yellow) for all readers, which suggests a comprehensive boost in ability to differentiate malignant IPNs. Notably, the most pronounced improvement was observed in reader 02 (the maximum area increment), while the least was observed in reader 12 (the minimum area increment).

### Confidence of readers with and without AI assistance

In the clinical trial, each IPN was graded by the readers on a scale of 1-10, with 1-5 being benign and 6-10 being malignant. **Extended Data Figure 3** and **4** show the changes in the number and percentage of benign or malignant nodules for each rating level before and after AI assistance. For nodules rated 1-5 by the twelve readers during independent reading, the proportions of true pathologically benign nodules were 84%, 83%, 71%, 67%, and 57%, respectively. For nodules rated 6-10, the proportions of pathologically malignant nodules were 57%, 66%, 74%, 78%, and 85%, respectively. With AI-assisted reading, the proportions of pathologically benign nodules among those rated 1-5 increased to 86%, 87%, 81%, 77%, and 66%, while the proportions of pathologically malignant nodules among those rated 6-10 were increased to 66%, 75%, 86%, 94%, and 97%, respectively. These results indicate that the AI model can substantially enhance the diagnostic accuracy of pulmonary nodules across all rating levels, with a pronounced impact on malignant nodule identification. This enhancement suggests that the model can increase diagnostic confidence for malignant cases while reducing false positives, thereby mitigating the risk of unnecessary radiation exposure or invasive interventions of patients.

Additionally, as shown in **Figure 5a**, the changes in diagnostic scores for each reader indicate that, after AI assistance, most readers assigned lower scores to benign nodules (mean decrease of 0.65 points) and higher scores to malignant nodules (mean increase of 0.25 points). When examining score changes at each level of unassisted scoring, nodules that were initially misclassified as benign (Score ≤ 5 without AI) showed an increase in mean scores after AI assistance, whereas those misclassified as malignant (Score > 5 without AI) demonstrated a decrease in mean scores following AI assistance. Similarly, both the **Extended Data Figures 5 and 6** show that with DeepFAN assistance, readers tended to assign lower malignancy scores to benign nodules and higher scores to malignant ones, reflecting increased diagnostic confidence and improved differentiation between benign and malignant nodules. Finally, with DeepFAN assistance, the overall kappa agreement coefficient increased from 0.285 to 0.417 ($P$=0.026) at the patient level, and from 0.313 to 0.421 ($P$=0.019) at the nodule level. The visualized correlation coefficient matrix shows that the agreement between each pair of readers was effectively improved with AI assistance (**Figure 5b&c**).

**Stratified analysis**

The diagnostic performance of DeepFAN, unassisted readers and AI-assisted readers was further analyzed and compared over subgroups stratified by patient characteristics including age and gender, nodule characteristics including diameter, density, location, and diagnostic difficulty, and reader characteristics including hospital affiliation, working experience and education level (highest academic degree). The diagnostic difficulty of a nodule was defined as low, intermediate, and high when more than two-thirds, between one-third and two-thirds, and less than one-third, respectively, of the unassisted readers correctly classified it.

Results from stratified analysis (**Supplementary Table 10 and 11**) show that DeepFAN exhibits significantly higher AUC compared to unassisted readers and AI-assisted readers over all subgroups while substantially improving the diagnostic ability of readers ($P$<0.001 for all AUCs). These results were consistent across the entire clinical trial dataset. In particular, over subgroups of nodules with low, intermediate, and high diagnostic difficulty, the AUC of DeepFAN reaches 0.994, 0.942, and 0.644 respectively, and DeepFAN enhances the diagnostic performance of readers in all difficulty levels (low: 0.913 to 0.956, △=0.043 [0.034-0.052]; intermediate: 0.447 to 0.642, △=0.195 [0.167-0.220]; and high: 0.113 to 0.229, △=0.116 [0.088-0.144]). These findings suggest that DeepFAN is robust to variations in nodule, patient and reader characteristics, maintaining competent predictive and stable assistive capabilities even for more challenging nodules.

From **Extended Data Figure 7**, we observe a steady enhancement in AUC when readers were assisted with AI. Meanwhile, nodules sized 20-30 mm ($P$<0.001), solid nodules ($P$<0.05), readers with Medical Doctor (M.D.) as their highest degree ($P$<0.001), and readers with clinical trial center I (Peking University People's Hospital) as their hospital affiliation ($P$<0.001) were associated with better diagnostic performance among subgroups respectively stratified by nodule diameter, nodule density, education level of readers, and hospital affiliation of readers across all three reading modes.

**Model visualization and interpretability**

**Figure 6** illustrates how DeepFAN utilizes chest CT to make decisions. Heatmaps were generated using the gradient-weighted class activation mapping (Grad-CAM) method[38], enhancing model interpretability at both the image and feature levels. Heatmaps help visualize and understand which areas of the feature map contribute most to a network output. Observations indicate that DeepFAN primarily relies on global features from the ViT module while incorporating local features from the CAL-ADL 3D ResNet, representing malignancy related nodule characteristics. Detailed descriptions of the two approaches used to capture local and global features are illustrated in the Methods section. Briefly, the ViT branch processes a 128×128×128 mm³ volume centered on the nodule, covering at least one-quarter of the lung and thereby allowing the capture of hidden nodule local features and contextual relationships (e.g., perinodular textures) and vascular connections. Meanwhile, the local feature branch captures nodule-specific details including density, lobulation, and spiculation. For example, DeepFAN successfully recognizes the smooth margin, solid density, and partial fat density of hamartoma (**Figure 6a**, benign nodule), as well as the irregular shape, heterogeneous density, lobulation, and spiculation associated with invasive adenocarcinoma (**Figure 6b**, malignant nodule). These features provide the AI model with a more comprehensive and clearer perspective for predicting malignancy versus benignity. Ultimately, GCN captures the relationships among these deep features (i.e. graph nodes), facilitating global and local information aggregation that improves diagnostic performance. However, some complex nodules were inherently confusing, exhibiting radiologic characteristics shared by both benign and malignant nodules. For example, epithelial hyperplasia (**Figure 6c**, benign nodule) shows dispersed morphology, heterogeneous ground-glass density, and multiple adjacent vascular branches, while invasive adenocarcinoma (**Figure 6d**, malignant nodule), that has an irregular shape, relatively high and uniform solid density, lobulation and long spiculation, can be easily confused with granulomatous inflammation. These shared radiologic characteristics and their combinations can interfere with AI's interpretation of a chest CT, which might ultimately result in a misclassification.

To delve deeper into the mechanisms of the proposed DeepFAN model for classifying IPNs as benign or malignant, logistic regression analyses were conducted to explore the correlation between DeepFAN's predictions and nodule characteristics (**Supplementary Table 12**). Multivariable analyses indicated significant associations between malignancy predictions and a subset of nodule characteristics, including large diameter (odds ratio [OR]=1.11, $P<0.001$), part-solid and ground-glass nodule density (OR=30.05, $P<0.001$ and OR=18.05, $P<0.001$), and presence of spiculation (OR=4.67, $P<0.001$) and lobulation (OR=4.26, $P<0.001$).

**Factors influencing Human-AI collaboration**

Sankey diagrams (**Extended Data Figure 8**) were used to demonstrate the flow of changes in diagnostic results from independent reading to AI-assisted reading for the twelve readers, and for reader 2 and reader 12 individually, as they were two special readers benefiting the most and least from the AI assistance according to the radar maps. Using the pathology reports as the reference standard, the false positive rate (FPR, defined in Method section) and false negative rate (FNR, defined in Method section) for DeepFAN were 15% and 5%, respectively, and the overall FPR and FNR for independent reading were 40% and 31%, respectively. With the assistance of the AI model, the readers corrected 43% of benign misdiagnoses and 41% of malignant misdiagnoses, reducing the FPR and FNR to 27% and 23%. Reader 2 had a higher misdiagnosis rate than reader 12 (34% vs. 27%). With the assistance of DeepFAN, reader 2 corrected 68% of misdiagnoses, while reader 12 only corrected 21%, leading to a more significant improvement in diagnostic accuracy for reader 2 compared to reader 12 (88% vs. 73% after DeepFAN assistance).

**Extended Data Figure 9** presents examples of human-AI collaboration outcomes using Grad-CAM, showcasing three distinct scenarios: (a) cases where AI correctly classifies nodules initially misclassified by radiologists, enabling correct revisions during collaboration; (b) cases where AI correctly classifies nodules misclassified by radiologists, but radiologists' predictions remain unchanged due to the nodules' strong misleading features; and (c) cases where both AI and radiologists, with or without AI assistance, misclassify nodules due to their highly deceptive imaging characteristics. These findings illustrate the AI's potential to improve diagnostic accuracy while also highlighting the challenges of human-AI collaboration in complex or ambiguous cases.

To further explore the factors affecting AI-assisted reading accuracy, 22 factors in five aspects, including diagnostic results, patient characteristics, nodule characteristics, CT image characteristics, and reader characteristics, were included for generalized linear mixed model analyses (**Supplementary Table 13**). In univariable analysis, malignant nodules, correct AI suggestions, correct independent readings, and larger nodule diameters are associated with higher AI-assisted reading accuracy ($\beta>0$, $P<0.05$), while interaction between correct AI suggestions and correct independent readings, nodules with ground-glass opacity, lobulation, higher diagnostic difficulty, and hospital affiliation and lower education level of the readers are associated with lower AI-assisted diagnostic accuracy ($\beta<0$, $P<0.05$). Subsequently, factors with $P<0.05$ were selected for further multivariable analysis. After adjustment for covariance, correctness of DeepFAN predictions ($\beta=1.72$, $P<0.001$), diagnostic difficulty of nodules (reference: low; intermediate, $\beta=-1.65$, $P<0.001$; high, $\beta=-2.68$, $P<0.001$), and education level of readers (reference: M.D.; M.M, $\beta=-0.53$, $P=0.034$; B.M., $\beta=-0.67$, $P=0.009$) were correlated with AI-assisted accuracy with statistical significance. The results of Grit score were shown in **Supplementary Table 14** and the translated reader questionnaire was provided in the **Supplementary Information**.

**Web-based AI platform**

To facilitate the promotion and application of DeepFAN in clinical practice, it has been implemented as a web-based AI platform (**Figure 7**). This platform enables user registration, case upload, and result generation. When patients complete a CT examination under the arrangement of clinicians, their information, including chest CT images, will be automatically uploaded to picture archiving and communication system (PACS). Radiologists can upload CT images they choose from PACS to this web-based AI platform. The platform will provide diagnostic advice generated from DeepFAN to radiologists, including the classification of IPNs as benign or malignant ones as well as the characteristics of nodules such as lobulation and spiculation. Finally, radiologists make the conclusion about the nature of pulmonary nodules refer to this information and provide reports to patients and clinicians.

**Discussion**

Although numerous studies have emerged on deep learning models for pulmonary nodule classification, AI diagnostic tools certified with rigorous MRMC clinical trials for clinical decision support remain scarce. In this study, we conducted the first officially registered multicenter, MRMC clinical trial with a paired study design in China (ChiCTR2400084624) to evaluate the performance of the proposed DeepFAN in assisting radiologists to differentiate malignant IPNs from benign ones in chest CT scans. By integrating global features and local nodule features, the proposed transformer-based DeepFAN model demonstrated excellent performance in IPN classification across multicenter test datasets representing various clinical scenarios, exhibiting strong robustness and generalization ability. Additionally, DeepFAN significantly improved the diagnostic accuracy, confidence and consistency of radiologists, showing its potential to homogenize healthcare services provided by medical practitioners with different educational backgrounds and working experience. Furthermore, studies on explainability revealed differences in assessment logics between human and AI, which provides valuable insights on clinical implementation pathways and future investigations in human-AI interaction.

Previous deep learning models have demonstrated potential in pulmonary nodule classification, frequently surpassing human experts in diagnostic accuracy[16-19,39-41]. However, most of them have been developed to operate on specific nodule types[19,40] (e.g., solid or part-solid) or prioritize local feature extraction while neglecting global contexts[17-19]. Only a few of them have been validated through rigorous clinical trials. To address these limitations, we present DeepFAN, an advanced deep neural architecture including a vision transformer for global context modeling, a CAL-ADL 3D ResNet for local feature extraction, and a graph convolutional network for hybrid feature fusion. DeepFAN was evaluated across four settings: (1) on an internal dataset from nine centers, it achieved high AUC and sensitivity, and further validated its core components; (2) on an external dataset (the clinical trial dataset) from three centers, it demonstrated satisfactory predictive capability by reporting high AUC, sensitivity, and specificity. (3) on the clinical trial dataset representing surgery-intended population, it outperformed 12 radiologists with 1–5 years of experience across various aspects; (4) on the NLST dataset representing western populations undergoing lung cancer screening, it demonstrated favorably robustness and generalization ability, achieving high AUC and near-perfect NPV. Notably, DeepFAN showed a higher NPV while a lower PPV on the NLST dataset compared to those on the clinical trial dataset. The discrepancy is reasonable as a larger proportion of benign nodules in the NLST dataset likely gives rise to an increased number of false positives (predicting a benign nodule as malignant), and even a small increase in the number of false positives can lead to a significant drop in PPV as the total number of positive predictions (denominator) increases while the number of true positives (numerator) remains small due to a decreased proportion of malignant nodules. Furthermore, DeepFAN offers interpretability because its internal information, such as GCN node weights and semantic characteristics linked to nodule malignancy, can be visualized to show the decision-making process. This interpretable framework underscores DeepFAN's superior diagnostic performance, surpassing both radiologists and previous state-of-the-art models[17,18,22].

To accurately evaluate the impact of AI-assisted diagnostic tools on clinicians' assessment of the malignancy risk of IPNs, we conducted a rigorously designed MRMC clinical trial to validate DeepFAN in real-world clinical practice. While previous studies highlighted considerable variability among radiologists in pulmonary nodule assessment[11,12,42-44], our clinical trial demonstrated substantial improvements in readers' diagnostic performance with increased accuracy across rating levels, enhanced confidence in nodule assessment, and a marked rise in inter-reader diagnostic consistency from fair to moderate. Although previous studies[30,45,46] also tried to leverage computer-aided tools to improve consistency or accuracy in pulmonary nodule diagnosis, limitations, such as absence of washout periods and absence of Asian populations, exist. Our findings provide robust and novel evidence that AI-assisted tools, exemplified by DeepFAN, can deliver consistent and accurate IPN assessments, which have the potential to alleviate patient anxiety, reduce unwarranted imaging, and mitigate radiation-related health risks.

The stratified analyses showed DeepFAN's robustness across various CT parameters (including different CT manufacturers and scanning/reconstruction protocols), patient cohorts and nodule characteristics. While the standalone DeepFAN model achieved robust and balanced diagnostic performance across most nodule subsets, its effectiveness was limited in high-difficulty nodules (AUC = 0.644), underscoring the need for further refinement, such as integrating multi-modal data or reinforcing training with high-difficulty datasets. Comparatively, the model showed greater improvement in intermediate-difficulty nodules (AUC improved from 0.447 to 0.642; $\Delta$=0.195) than in low-difficulty cases (from 0.913 to 0.956; $\Delta$=0.043). However, it provided limited benefit in high-difficulty cases (from 0.113 to 0.229; $\Delta$=0.116), highlighting the need for enhanced assistive strategies in challenging scenarios, such as improving AI interpretability. Despite these limitations, DeepFAN consistently provided reliable assistance to readers across diverse patient cohorts (encompassing different ages and genders), nodule attributes (such as size, density and location), and reader characteristics (including education, experience, and clinical center) within the clinical trial dataset.

After thorough correlation analysis, we discovered that AI-assisted reading accuracy was closely associated with the correctness of DeepFAN's predictions, the diagnostic difficulty of nodules, and the education level of readers after adjustment for the covariance. Notably, readers holding M.D. degrees demonstrated superior diagnostic performance (P<0.001). And all readers at Peking University People's Hospital held M.D. degrees, resulting in the highest average accuracy in the study (P<0.001). This fact likely reflects systematic disparities in resource allocation and professional development opportunities as larger cities like Beijing, with larger populations (21.9 million vs. 12.4 million vs. 2.5 million, Beijing vs. Wuhan vs. Huangshi according to the national[47] and Hubei province Bureau of Statistics[47,48]) and better infrastructure, attract students pursuing advanced degrees and expose junior radiologists to richer diagnostic experience[48,49]. Importantly, while DeepFAN allowed those with B.M. and M.M. degrees to outperform M.D. readers without AI assistance in terms of AUC, sensitivity and specificity (**Supplementary Table 10**), a notable performance gap remained between M.D. and non-M.D. readers when both groups used DeepFAN. This suggests that while DeepFAN improves the performance of non-M.D. readers, it does not eliminate expertise-related disparities. Despite these, DeepFAN has shown potential to assist a broad spectrum of medical professionals—including those without M.D. degrees—by improving diagnostic accuracy across diverse levels of training.

Although DeepFAN demonstrated substantial improvements in readers' diagnostic accuracy during the clinical trial, AI-assisted reader performance did not exceed that of the AI model itself. Further analysis revealed that this limitation was unrelated to a reader's experience, personality, or attitude toward AI but was instead influenced by deceptive nodule characteristics[32], such as benign nodules with malignant-looking radiological features. Such cases often lead readers to reject correct AI suggestions due to entrenched diagnostic patterns. DeepFAN's ability to transcend traditional radiological features by incorporating global patterns underscores its advanced diagnostic capabilities while highlighting the need for AI tools to reshape conventional diagnostic approaches[50,51]. Despite these challenges, DeepFAN enabled readers to correct 43% of benign and 41% of malignant misdiagnoses, reducing false positive and negative rates to 27% (vs. 40% without DeepFAN) and 23% (vs. 31% without DeepFAN), respectively. Even the least improved reader corrected 21% of errors while the most improved reader corrected 68%, reflecting DeepFAN's adaptability across diverse user characteristics.

The rationale behind our choice of radiologists with less than five years of experience is threefold. First, in frontline hospitals in smaller cities, initial diagnoses are often made by less experienced radiologists due to an imbalanced distribution of medical resource. This presents an opportunity for AI to bridge expertise gaps, advancing the goal of equitable healthcare. Second, China's two-tier reading system, where junior radiologists perform initial assessment followed by review performed by senior radiologists, could benefit significantly from improved diagnostic accuracy at the junior level, alleviating senior radiologists' workload and improving review efficiency. Third, previous breast cancer diagnostic research has demonstrated AI's potential to replace junior doctors in initial readings, aligning this study with the broader trends of optimizing clinical workflows through AI integration[52]. While DeepFAN has shown promise in reducing diagnostic discrepancies among radiologists with varying educational backgrounds (BM vs. MD), it remains unclear whether AI-assisted junior radiologists can match or exceed the diagnostic performance of experienced senior radiologists. Further research is needed to evaluate DeepFAN's applicability and effectiveness in more complex clinical settings involving senior radiologists.

Our DeepFAN model has shown promising results, but limitations exist. First, all the test datasets were retrieved retrospectively, and a prospective clinical study warrants more robust evidence. Nonetheless, the results from our multicenter clinical trial and on the NLST dataset have demonstrated the model's generalization ability. Second, only pathologically confirmed lung nodules were included in our clinical trial, this criterion may favor the inclusion of more suspicious nodules, which could impact the generalizability of our findings. To mitigate potential selection bias, we validated the model's performance in a stratified subgroup of lung nodules measuring 4–10 mm and in the GGN subgroup, and also conducted testing in the NLST lung cancer screening dataset. Third, the evaluation of DeepFAN's efficacy in AI-assisted reading was limited to inexperienced radiologists, who represent frontline practitioners for reporting potentially malignant pulmonary nodules. Further investigation is needed to assess its utility among experienced radiologists, such as how senior radiologists interact with AI predictions and whether AI can reduce inter-observer variability in multi-tiered diagnostic system, and its effectiveness across diverse clinical settings, including those in Western healthcare systems. Last, DeepFAN only provided binary classification results to human readers during the AI-assisted reading sessions in our clinical trial. It remains to be explored whether providing additional information, such as a ternary classification including an extra category for nodules with uncertain diagnoses or the premalignant category, would increase the acceptance of AI suggestions by radiologists.

In conclusion, we have conducted the first MRMC clinical trial in China to evaluate the efficacy of our developed DeepFAN model in assisting radiologists to assess the malignancy risk of IPNs in chest CT scans. DeepFAN has not only exhibited outstanding and robust diagnostic performance on IPNs across multiple internal and external datasets as well as diverse clinical scenarios, but also consistently enhanced the diagnostic accuracy, confidence and inter-reader consistency of junior radiologists via human-AI collaboration. By facilitating more uniform healthcare services across regions, DeepFAN holds the potential to mitigate IPN-related anxiety and reduce unnecessary imaging surveillance, thereby lowering radiation-related health risks and avoiding excessive medical expenditures.

## Methods

### Ethics approval

This study includes a retrospective, multi-center, multi-reader multi-case (MRMC) clinical trial with a paired design (Chinese Clinical Trial Registry: ChiCTR2400084624). It was conducted in accordance with the Declaration of Helsinki. The Institutional Review Boards of the primary sponsor (Peking Union Medical College Hospital, JS-2805) and the three enrollment and research sites (Peking University People's Hospital, 2022PHA118-001; Wuhan Third Hospital, Wuhan No.3 QX2023-002; and Huangshi Central Hospital, Lun Kuai Shen [2023] No. 2) approved the trial and waived informed consent for patients. Informed consents for twelve readers were obtained prior to the study.

**Clinical trial dataset**

The officially registered national clinical trial was designed to not only evaluate the performance of DeepFAN in distinguishing malignant pulmonary nodules from benign ones but also examine its efficacy in assisting junior radiologists under real clinical settings.

The clinical trial retrospectively collected demographic data and unenhanced chest CT scans from 400 consecutive patients across the three clinical trial centers: Peking University People's Hospital (center I), from September 2022 to December 2022; Wuhan Third Hospital (center II), from August 2020 to February 2023; and Huangshi Central Hospital (center III), from September 2020 to November 2022. Among the three centers, Peking University People's Hospital had a significantly higher surgical volume compared to the other two centers. To ensure relatively balanced representation and distribution of the 400 cases across centers, we extended the data collection period for the lower-volume centers while maintaining overlapping periods (e.g., including 2022 data from all centers).

The inclusion criteria are as follows (**Extended Data Figure 1**): (a) patient age ≥ 18 years; (b) availability of postoperative pathology; (c) nodule diameter ≥ 4mm and ≤ 3cm; (d) interval between the latest preoperative CT scan and surgery ≤ one month; (e) CT slice thickness < 2mm and reconstruction slice increment ≤ slice thickness; and (f) DICOM-compliant CT images. The exclusion criteria are as follows: (a) nodules reported by pathology could not be accurately located in CT images; (b) incomplete CT scan range, poor CT image quality (metal or breathing artefacts), or postoperative CT; (c) metastatic lesions; and (d) redundant malignant cases were excluded to ensure a malignant-to-benign ratio close to 1.

Baseline characteristics of the enrolled patients and CT protocols are shown in Table 1 and Supplementary Tables 6-7. All data were de-identified prior to model training, validation, testing and clinical trials.

**Clinical trial design**

The study adopted a fully crossed MRMC design (**Figure 2**), which included twelve readers from three clinical trial centers: readers 01–03 from Peking University People's Hospital, readers 04–07 from Wuhan Third Hospital, and readers 08–12 from Huangshi Central Hospital. All readers had 1–5 years of working experience in general diagnostic imaging of CT scans, with an average of 2.83 years. To avoid selection bias, all readers were randomly selected from the pool of eligible radiologists. Detailed information about the readers can be found in **Supplementary Table 8**.

The MRMC study comprised two rounds of image reading, separated by a four-week washout period. In the first round, Group A served as the control group, independently assessing the benignity and malignancy of the 400 cases, while Group B served as the test group, using the AI model for assistance (all 400 cases). In the second round, the roles were reversed, with Group A acting as the test group and Group B as the control group. The study platform guided readers to outline a nodule in the CT scan of a trial case with a rectangle box. If AI assistance was enabled, the platform displayed a card with the AI-predicted classification of the nodule (benign or malignant); otherwise, the card remained blank. To ensure accurate nodule localization, each reader was provided with a handbook showing the locations of the 463 nodules defined in the ground truth in the original CT scans. The readers were then instructed to annotate the nodules with tight rectangular boxes. For independent reading, the readers assigned a binary label (benign or malignant) and a risk score (1-5 for benign; 6-10 for malignant) to every located nodule. For AI-assisted reading, the AI model automatically provided diagnostic results for the nodules annotated by the readers, who then referred to these AI-generated results before providing their final labels and scores. No time constraints were imposed on cases reading.

Throughout the clinical trial, all readers: (a) were blinded to clinical and pathological information of the trial cases; (b) were unaware of the benign-to-malignant ratio of pulmonary nodules in the clinical trial dataset; (c) were not informed of the AI model's diagnostic performance; (d) did not recall results from the first round during the second round; and (e) received no training beyond that required for using the AI software.

**Ground truth**

To ensure that the included cases were qualified, an expert group comprising experienced thoracic surgeons (11-30 years of experience), pathologists (15-23 years of experience), and radiologists (20-26 years of experience) from the three centers were formed. The expert group was responsible for finalizing case selection and providing ground truths for each case. Specifically, thoracic surgeons were responsible for screening cases according to the inclusion and exclusion criteria, and correlating nodules in surgical reports with pathological findings. Pathologists reviewed the pathology reports while radiologists delineated target nodules in CT images. The malignancy of pulmonary nodules was determined according to the ICD-10 (2015 edition, accessible via https://icd.who.int/browse10/2015/en).

**Evaluation metrics**

The primary evaluation metrics were ROC-AUC, accuracy, sensitivity, specificity at the nodule level. The test group and the control group provided potential determinations of benign or malignant pulmonary nodules based on the specific characteristics of the nodules, along with a confidence score indicating the likelihood of benign or malignant nature (ranging from 1 to 10, where 1-5 indicates benign and 6-10 indicates malignant). Based on the confidence scores at the nodule level, the ROC-AUC for the determination of benign or malignant pulmonary nodules was calculated for both the test group and the control group, and a comparison was made between the groups.

   The secondary evaluation metrics were positive predictive value (PPV), negative predictive value (NPV), F1-score, false positive rate (FPR) and false negative rate (FNR) at the nodule level. For sensitivity analysis, ROC-AUC, sensitivity, specificity, and accuracy at the patient level were calculated. At the patient level, the confidence score was defined as the highest score among all nodules for a single patient. A positive result at the patient level was defined as having at least one positive/malignant nodule, while a negative result at the patient level was defined as having no positive/malignant nodules. The formulas for the above-mentioned metrics were defined as follows:

**Sensitivity** = TP / (TP + FN)

**Specificity** = TN / (FP + TN)

**Accuracy** = (TP + TN) / (TP + FP + TN + FN)

**PPV** = TP / (TP + FP)

**NPV** = TN / (TN + FN)

**F1-score** = 2×TP / (2×TP + FP + FN)

**FPR** = FP / (FP + TN)

**FNR** = FN / (FN + TP)

FP, TP, FN, and TN stand for false positives, true positives, false negatives, and true negatives, respectively.

**Model development cohort and generalization test cohort**

CT images of pulmonary nodules taken between November 2011 and October 2021 were retrospectively collected from nine hospitals across seven provinces in China, and randomly divided into training, validation, and internal test sets in a ratio of 7:1:2 (no overlap among these sets at the patient level), resulting in a training set of 5,636 patients with 7,873 nodules, a validation set of 831 patients with 1,216 nodules and an internal test set of 1,705 patients with 2,349 nodules. The benign and malignant nodules in the above datasets were all pathologically confirmed.

   Considering the potential bias in the proportion of benign and malignant nodules caused by the inclusion of pathologically confirmed nodules only, we further evaluated model performance on the National Lung Screening Trial (NLST) dataset that includes benign nodules confirmed through patient follow-up.

   The inclusion and exclusion criteria for the development, validation and internal test set cohort were as follows:

(a) patient age ≥ 18 years; (b) lung nodules detected in unenhanced chest CT were pathologically confirmed and could be accurately localized in CT images through pathology reports and/or surgical records; (c) CT examination was performed before surgery, and slice thickness was less than 2mm; (d) DICOM-compliant CT images; and (e) exclusion of metastatic tumors ; (f) exclusion of CT images with poor quality affecting doctors' diagnosis or incomplete depiction of nodules.

The inclusion and exclusion criteria for the NLST dataset were as follows: (a) patients had undergone chest CT examination and non-calcified nodule/mass (diameter, ≥4mm and ≤30mm) was found on any screen; (b) benign and malignant cases confirmed by pathology or follow-up; (c) DICOM-compliant CT images; and (d)exclusion of CT images with slice thickness greater than 2mm.

For developmental dataset, each pulmonary nodule was manually annotated by radiologists with more than five years of experience in chest CT diagnostic imaging from a tertiary A-level hospital. The reviewing doctors were radiologists with more than ten years of experience in chest CT diagnostic imaging, who also worked in a tertiary A-level hospital. For the NLST dataset, we followed the methodology from Venkadesh et al.'s study[17] to locate benign and malignant nodules. Briefly, for participants diagnosed with lung cancer, a board-certified radiologist (J.W.) retrospectively reviewed all available imaging data across the screening periods to accurately identify malignant nodules within the tumor-affected lobe. For non-cancer participants, two trained medical students (L.P. & Z.Y.) independently reviewed the CT images using the NLST-provided lobar locations and CT section numbers to locate nodules. Inter-reader discrepancies were resolved by a senior radiologist (L.S.). Nodules with an average diameter < 4 mm were excluded in accordance with established size criteria.

**Data preprocessing**

Data preprocessing was conducted before model training. Due to variations in pixel spacing and CT slice thickness, the CT images were linearly interpolated into 3-dimentional isotropic images, with voxel spacing set to 0.6 mm × 0.6 mm × 0.6 mm. Image patches enclosing nodules were cropped from the original CT images and used as training samples. To include rich contextual information around a nodule, the cropped image patch is centered at the nodule, but large enough to cover its surrounding area. In this study, image patches with a size of 128 × 128 × 128 pixels were used. Subsequently, conventional data augmentation techniques, such as random cropping and flipping, were applied. The augmented image patches were then used for model training.

**Architecture of DeepFAN**

The neural architecture of DeepFAN consists of three primary modules: a vision transformer (ViT) module for global nodule feature extraction, a fine-grained module for local feature and attribute feature extraction, and a graph convolutional network (GCN) for feature fusion (**Figure 1**). Details of the model construction are as follows.

**a) Data preprocessing.** The data preprocessing pipeline involves resampling CT scans into uniform isotropic voxels (0.6 mm × 0.6 mm × 0.6 mm), extracting 128-pixel cubic patches centered at nodules to include contextual information, and applying data augmentation (such as random cropping and flipping) before model training.

**b) ViT module.** We further subdivide every 128 × 128 × 128 training sample into eight 64 × 64 × 64 3D patches. Each 3D patch passes through a patch embedding block implemented as a stride-2 convolutional layer followed by a down-sampling layer. Patch embedding transforms a 3D patch into a 4096-dimensional patch token. The ViT module also has a class token—a trainable vector prepended to the sequence of patch tokens. The class token aggregates global information from all tokens during the learning process, serving as the model's output for classification tasks. Learnable 1D positional embeddings are added to every token to enhance positional awareness. Next, one class token and eight patch tokens are fed into twelve transformer blocks, which generate nine corresponding output feature vectors, including one for the class token and eight for the patch tokens. Then the class feature vector passes through an FC layer to predict the probability of malignancy. At the same time, it is also fed into another FC layer that produces a 64-dimensional feature vector ($H_T^0$). As to the other eight feature vectors, each of them passes through a separate FC layer, generating a 64-dimensional feature vector (denoted as $H_T^1$ to $H_T^8$), which captures global characteristics of the input sample. Thus, nine feature vectors are fed into the GCN module as input nodes.

Notably, the term "global" in this context carries dual implications. First, the self-attention mechanism ingrained within the ViT model inherently computes pairwise attention among the input tokens, thereby capturing comprehensive information from the entire input. Second, the term also signifies that the input sample is sufficiently expansive to encompass the nodule itself along with its immediate surroundings.

**c) Fine-grained model.** The input has 128 × 128 × 128 pixels in this module. After passing through three ResNet[53] stages (each containing 6, 9, 12 blocks), the resulting feature map is denoted as $F \in R^{512 \times 16 \times 16 \times 16}$, where 512 represents the number of channels and the spatial dimensions of the feature map are 16 × 16 × 16. Then, we integrate an Attention Dropout Layer (ADL)[54] with counterfactual causal learning (CAL)[55] to further refine feature map $F$. Specifically, $F$ is processed with ADL to obtain an attention feature map $A$. Meanwhile, randomized attention mechanisms are adopted to generate a counterfactual intervention feature map $\bar{A}$. Next, feature maps $F$ and $A$ are fed into a bilinear attention pooling layer (BAP)[56]. The output is flattened and then passed through three parallel FC layers, resulting in three 64-dimensional feature vectors ($H_C^0, H_C^1$ and $H_C^2$). Feature maps $F$ and $\bar{A}$ are processed similarly to produce another three 64-dimensional feature vectors ( $\bar{H}_C^0, \bar{H}_C^1$ and $\bar{H}_C^2$). Afterwards, features $H_C^0$ and $\bar{H}_C^0$ are passed through two independent FC layers, and the element-wise difference between the results,is used to predict lobulation. Features $H_C^1$ and $\bar{H}_C^1$ are processed similarly to predict spiculation. Likewise, features $H_C^2$ and $\bar{H}_C^2$ are also processed similarly to classify the nodule in the input sample into three density categories: ground-glass nodule, solid nodule and part-solid nodule.

Note that lobulation, spiculation and density are three important radiological signs of a lung nodule. To pass information from the fine-grained module to the GCN, an element-wise subtraction is performed between $H_C^0$ and $\bar{H}_C^0$ yielding a feature vector $H_{C0}$, between $H_C^1$ and $\bar{H}_C^1$ yielding a second feature vector $H_{C1}$, and between $H_C^2$ and $\bar{H}_C^2$ yielding a third feature vector $H_{C2}$. $H_{C0}$, $H_{C1}$, and $H_{C2}$ respectively represent the lobulation, spiculation and density features of the lung nodule, and serve as three additional input nodes of the GCN module.

**d) GCN model.** The GCN module leverages and fuses a comprehensive set of multi-scale features. Its core structure follows the design outlined in the article by Zhao et.al.[26], which incorporates three layers of residual graph convolution[57] and dynamically updates the edge weights between node features. The nine feature vectors ($H_T^0$ to $H_T^8$) generated by the ViT module and the three feature vectors ($H_{C0}$, $H_{C1}$ and $H_{C2}$) produced by the fine-grained module are placed side by side to form a matrix $H_{all} \in R^{12 \times 64}$. This matrix is then fed into the first layer of the GCN. Here, 12 represents the number of nodes, and 64 indicates the feature dimension of each node. At the end, we use an FC layer to obtain the final probability of malignancy.

**e) Loss function.** During the training process, we enable deep supervisions within individual modules of our network architecture. Thus, the final loss function comprises multiple terms as follows,

$$Loss_{TCG} = \omega_0 L_T^0 + \omega_1(L_{c0} + L_{c1} + L_{c2}) + \omega_2 L_{all} + \omega_3 L_G$$

where

$L_T^0$ is the loss term for the benign and malignant binary classification in the ViT module;

$L_{c0}, L_{c1}, L_{c2}$ are the loss terms for predicting the probabilities of lobulation, spiculation and density in the fine-grained module, respectively;

$L_{all}$ is the loss term for the benign and malignant binary classification using the feature matrix $H_{all}$ followed by an FC layer;

$L_G$ is the loss term for the benign and malignant binary classification in the GCN module.

In this study, $\omega_0$ was set to 0.2, $\omega_1$ to 0.2, $\omega_2$ to 0.2 and $\omega_3$ to 0.4. All loss terms have a cross-entropy form.

**f) Model training.** All models were trained using the PyTorch[58] deep learning framework. The optimization method used during training was Adaptive Moment Estimation (Adam). Training was conducted on four NVIDIA GeForce RTX 3090 GPUs.

During training, we first respectively optimized the ViT module and the fine-grained module independently until parameters stabilized. Both modules were trained for 1200 epochs with initial learning rates of 0.0002 and 0.01, respectively. The learning rate was reduced to 10% of its previous rate at the 400th and 800th epochs. Next, we optimized the GCN module while freezing the parameters of the other two modules. The GCN module was trained for 200 epochs with an initial learning rate of 0.01. and the learning rate was reduced to 10% of its previous rate at the 80th and 160th epochs. Finally, we optimized the final loss function $Loss_{TCG}$ by fine-tuning all the parameters of the ViT, fine-grained, and GCN modules. This phase involved training for 1400 epochs with an initial learning rate of 0.00001. The learning rate was reduced to 10% of its previous rate at the 300th, 600th, and 900th epochs. Model parameters were saved every 30 epochs, and then all saved checkpoints were tested on the validation set. The model with the highest AUC on the validation set was selected and tested on the testing set.

**Statistics and reproducibility**

The sample size was determined with the superiority test of AUROC via the multi-reader diagnostic test research software developed by Kevin M. Schartz and Stephen L. Hillis[59] as detailed in sample size estimation in the **Supplementary Information**. This was a single-blind, multi-reader diagnostic study. Readers were blinded to all clinical, pathological, and outcome information as well as to the AI model's diagnostic performance. Data analysis was performed independently by investigators who were not involved in image interpretation. The diagnostic performance of the AI model or readers was assessed using AUC, sensitivity, specificity, accuracy, PPV, NPV, and F1-score, with 95% CI. The 95% CI was calculated through the nonparametric bootstrap method with 1,000 resampling events. The threshold for DeepFAN to determine malignancy was set to the value that optimizes the model's performance (F1-score) over the validation set. This threshold was consistently applied across all test sets (internal test set and NLST test set) and clinical trial dataset. The ROC curves for readers (with and without DeepFAN assistance) were generated according to the malignancy scores they assigned to the nodules, while sensitivity and specificity were calculated using the binary labels provided by the readers. Interobserver diagnostic agreement was measured with Cohen's kappa. A logistic regression model was used to analyze the relationship between nodule characteristics and AI malignancy predictions. The generalized linear mixed model was used to explore factors influencing the accuracy of human-AI collaboration, with readers and cases as random effects and other factors as fixed effects.

Appropriate statistical tests were used to analyze data, as described in each figure and table legend. Normality was evaluated by combining statistical tests with graphical methods, accompanied by an assessment of homogeneity of variances as appropriate. Unless otherwise specified, all tests were two-sided, and $P$ values less than 0.05 were considered statistically significant. The Bonferroni method was applied to adjust for multiple comparisons. Statistical analyses were performed using R (version 4.4.1) and Python (version 3.6.13). Data preprocessing and model development were conducted using the PyTorch (version 1.10.0) deep learning framework. No reader-evaluation data were excluded from any analysis. During the MRMC clinical trial, all readers were randomly selected from the pool of eligible radiologists and were subsequently randomized in an even manner to either the control or test group. Please refer to the "**Clinical trial design**" section for more details.

**Data availability**

Proprietary training and clinical trial datasets are unavailable due to ethical and research restrictions. The NLST dataset (Radiology CT Images and Clinical data including data dictionaries) is available at https://www.cancerimagingarchive.net/collection/nlst/. Additionally, the Patient IDs of the NLST subset analyzed in this study is available on GitHub at https://github.com/zhjtwx/DeepFAN/blob/main/sample_data/Filtered_NLST_Subset.csv. Other supporting data are available from the corresponding authors (Lan Song, songl@pumch.cn; Yizhou Yu, yizhouy@acm.org) upon request, subject to institutional approval and a 6-week review. Source data are provided with this paper.

**Code availability**

The code for DeepFAN is available on Github (**https://github.com/zhjtwx/DeepFAN**).

**Acknowledgements**

Z.J. is funded by the Scientific and Technological Innovation 2030-New Generation Artificial Intelligence Project of the National Key Research and Development Program of China (2020AAA0109503) and the Beijing Municipal Science and Technology Program (Project No. Z201100005620008). L.S. is funded by the National Natural Science Foundation of China (82171934) and Peking Union Medical College Hospital Talent Cultivation Program Category C (UBJ10148). Y.Y. is funded by Hong Kong Research Grants Council through General Research Fund (Grant 17207722). The authors thank the National Cancer Institute for access to NCI's data collected by the National Lung Screening Trial. The funders had no role in study design, data collection and analysis, decision to publish or preparation of the manuscript. Additionally, we thank the radiologists, the investigators and research coordinators involved in the trial.

**Author Contributions Statement**

Y.Y., L.S., N.H., Z.J., and Z.Zhou contributed to study design and supervision. Z.Zhu, G.H., W.T., K.G, Y.Y., and Z.Zhou contributed to data analysis, model development and writing of the manuscript. K.X., X.Q., and Y.T. contributed to data acquisition. L.S., C.S., J.W., Z.Y., L.P. and W.S. contributed to data analysis and interpretation. W.H. and M.S. contributed to statistical verification. All authors critically revised and approved the manuscript.

**Competing Interests Statement**

W.T., K.G. and Z.Zhou are full-time employees of the AI laboratory at Deepwise Healthcare. Z. Zhou participated as a technical consultant, contributing to the MRMC study design. The DeepFAN system used in this study was developed by Beijing Deepwise & League of PHD Technology Co., Ltd. Prior to this publication, DeepFAN had obtained Class III certification from the National Medical Products Administration (NMPA) of China (Approval No. 20243211932). These authors did not represent a conflict of interest with respect to the execution of this study or the interpretation of data presented in this report. All other co-authors declare no competing interests.

**Table 1. Baseline characteristics of patients and pulmonary nodules in datasets from three clinical trial centers**

| Variable name | Clinical trial center I | Clinical trial center II | Clinical trial center III | *P* value |
|---|---|---|---|---|
| **Patient characteristics** | | | | |
| Total no. of patients | 166 | 46 | 188 | |
| Age (year)* | 59 ± 10 | 60 ± 10 | 59 ± 10 | 0.582 |
| Sex | | | | 0.457 |
|    Male | 78 (47) | 26 (57) | 97 (52) | |
|    Female | 88 (53) | 20 (43) | 91 (48) | |
| Nodule type | | | | 0.003^ |
|    Single | 139 (84) | 46 (100) | 167 (89) | |
|    Multiple# | 27 (16) | 0 (0) | 21 (11) | |
| **Nodule characteristics** | | | | |
| Total no. of nodules | 204 | 46 | 213 | |
| Nodule diameter (mm)* | 11.2 ± 5.6 | 15.6 ± 5.6 | 14.0 ± 5.6 | <0.001 |
| Nodule density | | | | <0.001 |
|    SN | 63 (31) | 26 (57) | 103 (48) | |
|    PSN | 76 (37) | 13 (28) | 79 (37) | |
|    GGN | 65 (32) | 7 (15) | 31 (15) | |
| Nodule location | | | | 0.699 |
|    RUL | 52 (25) | 16 (35) | 57 (27) | |
|    RML | 18 (9) | 1 (2) | 17 (8) | |
|    RLL | 57 (28) | 10 (22) | 49 (23) | |
|    LUL | 42 (21) | 11 (24) | 46 (21) | |
|    LLL | 35 (17) | 8 (17) | 44 (21) | |
| Spiculation | | | | <0.001 |
|    No | 125 (61) | 15 (33) | 68 (32) | |
|    Yes | 79 (39) | 31 (67) | 145 (68) | |
| Lobulation | | | | 0.005 |
|    No | 43 (21) | 3 (7) | 24 (11) | |
|    Yes | 161 (79) | 43 (93) | 189 (89) | |
| Pathology | | | | 0.269 |
|    Benign | 98 (48) | 27 (59) | 97 (46) | |
|    Malignant | 106 (52) | 19 (41) | 116 (54) | |

Unless otherwise indicated, numbers here are counts or percentages (in parentheses).

*Data are mean ± standard deviation. #Number of nodules per case ranges from two to seven. ^*P* values were obtained using Fisher's exact test, while all other categorical variables were compared with chi-square tests. *P* values for age and nodule diameter were analyzed using Kruskal-Wallis tests. All statistical tests were two-sided. *Abbreviations:* SN = solid nodule, PSN = part-solid nodule, GGN = ground-glass nodule, RUL = right upper lobe, RML = right middle lobe, RLL = right lower lobe, LUL = left upper lobe, LLL = left lower lobe.

**Figure Legends**

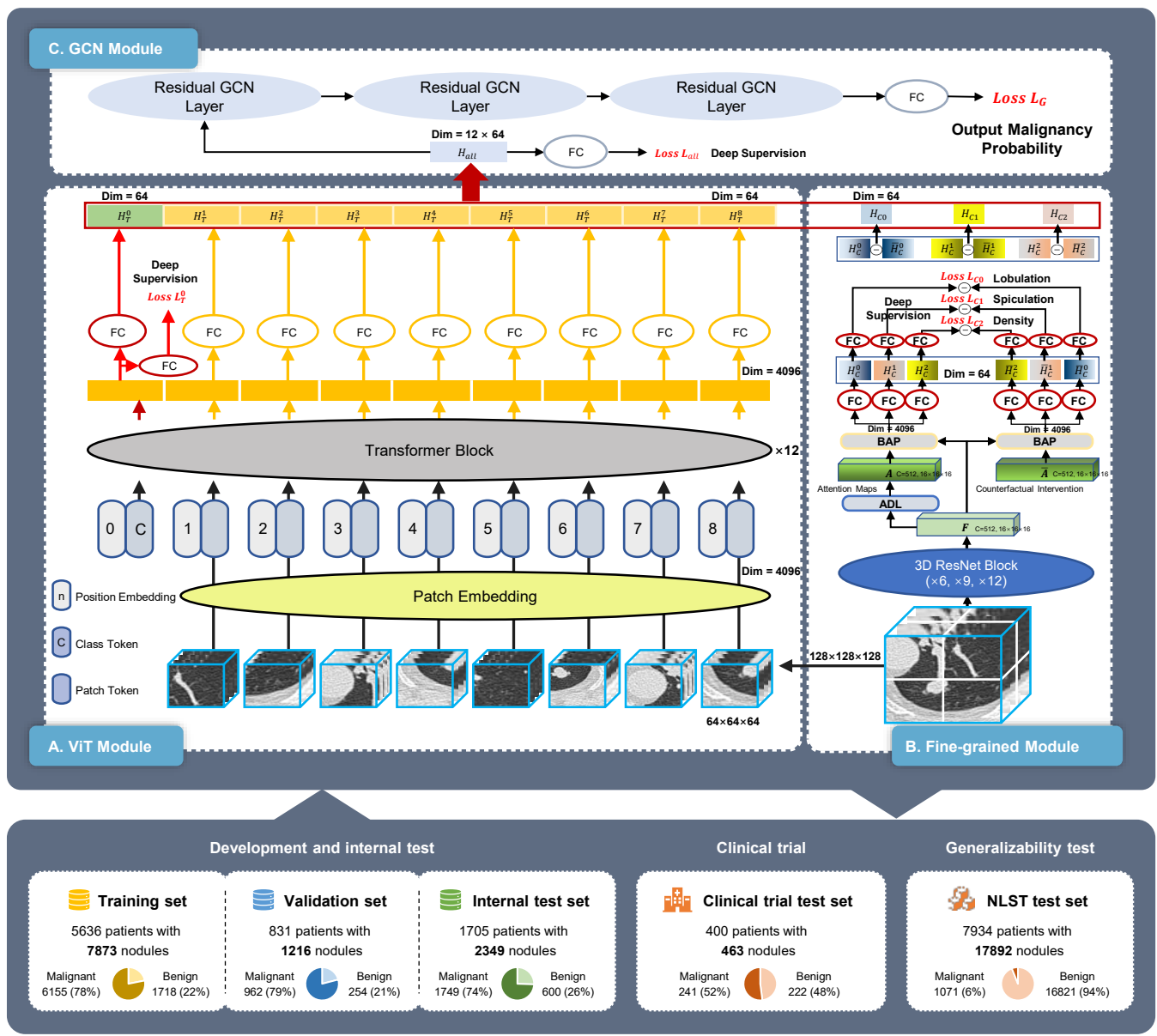

**Figure 1. Neural architecture of DeepFAN.**

DeepFAN integrates three modules. The first module utilizes a ViT to effectively capture global features of pulmonary nodules and their surroundings, representing the overall morphology and distribution characteristics of the nodules. The second module employs a fine-grained CAL-ADL 3D ResNet, which extracts features representing detailed radiologic characteristics of the nodules. In the last step, a GCN is introduced to learn the relationships between the global features extracted using ViT and the local features extracted using CAL-ADL 3D ResNet, aiming to comprehensively understand the correlation between pathological classification and pulmonary nodule characterization. DeepFAN was developed using a training set of 5,636 patients and the hyperparameters of the DeepFAN architecture were tuned on a validation set of 831 patients. The performance of DeepFAN was evaluated on an independent internal test set of 1,705 patients, a multicenter clinical trial test set of 400 patients and the NLST dataset of 7,934 patients. Boldface numbers represent numbers of nodules.

*Abbreviations:* ViT = vision transformer, CAL = counterfactual attention learning, ADL = attention dropout layer, 3D = three-dimensional, ResNet = residual network, GCN = graph convolution network, NLST = national lung screening trial, BAP = bilinear attention pooling, FC = fully connected, Dim = dimensions, C = channel, ⊖ : subtract.

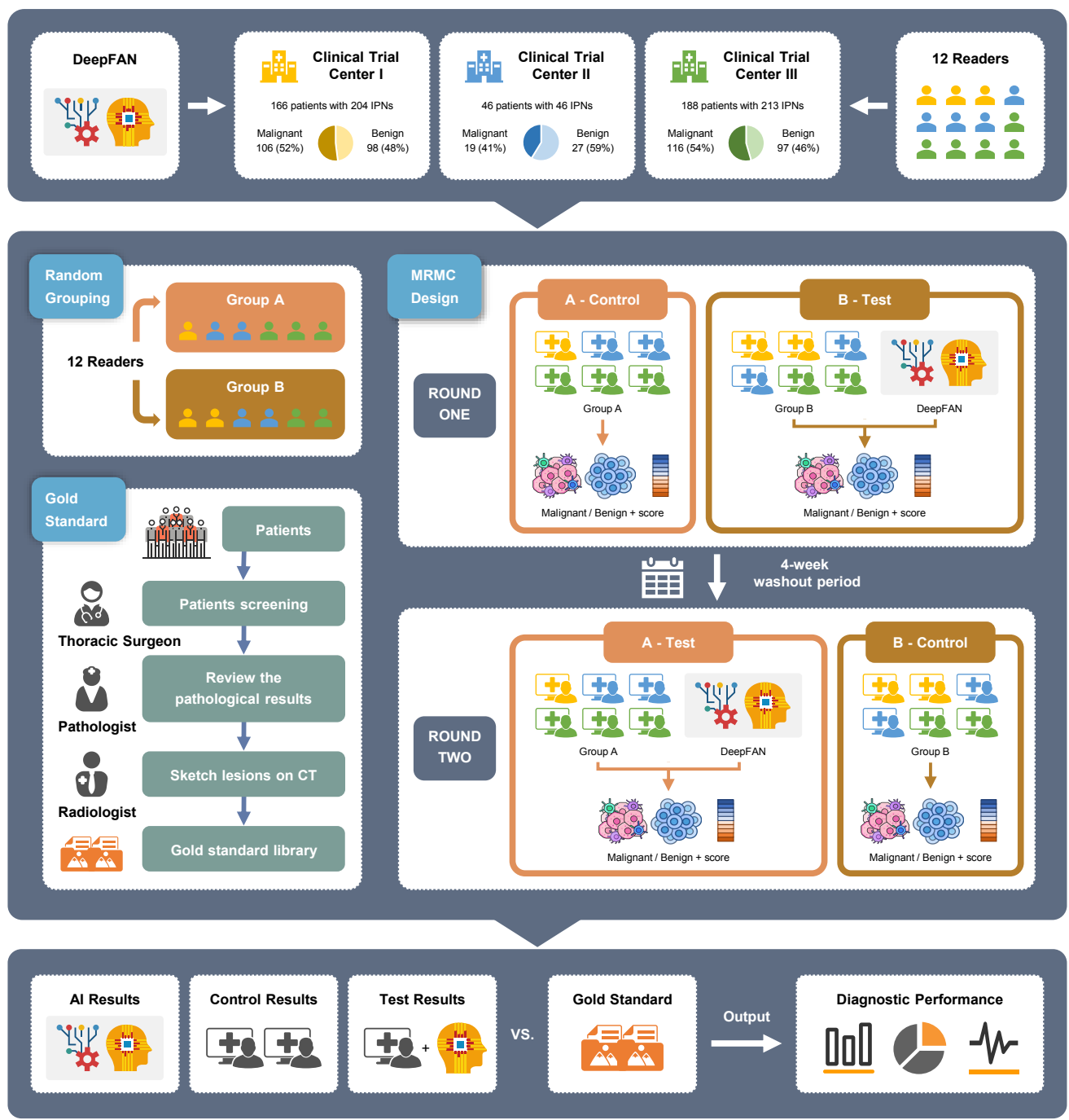

**Figure 2. Clinical trial workflow.**

The study included three retrospective datasets (400 patients with 463 IPNs) from hospitals with national clinical trial qualifications. Twelve readers selected from the same three institutions were randomly assigned to group A and B for paired CT image assessment. The MRMC procedure consisted of two reading rounds, separated by a 4-week washout period. In the first phase, group A served as the control group, assessing the benignity and malignancy of the cases without AI, while Group B served as the test group, using AI (DeepFAN) to assist assessment. In the second phase, the roles were reversed, with group A acting as the test group and group B as the control group. Based on a gold standard reference library developed collaboratively by thoracic surgeons, pathologists, and radiologists, the diagnostic performance of the AI model, individual readers, and AI-assisted readers were analyzed and compared.

*Abbreviations:* AI = artificial intelligence, IPN = incidental pulmonary nodule, CT = computed tomography, MRMC = multireader multicase.

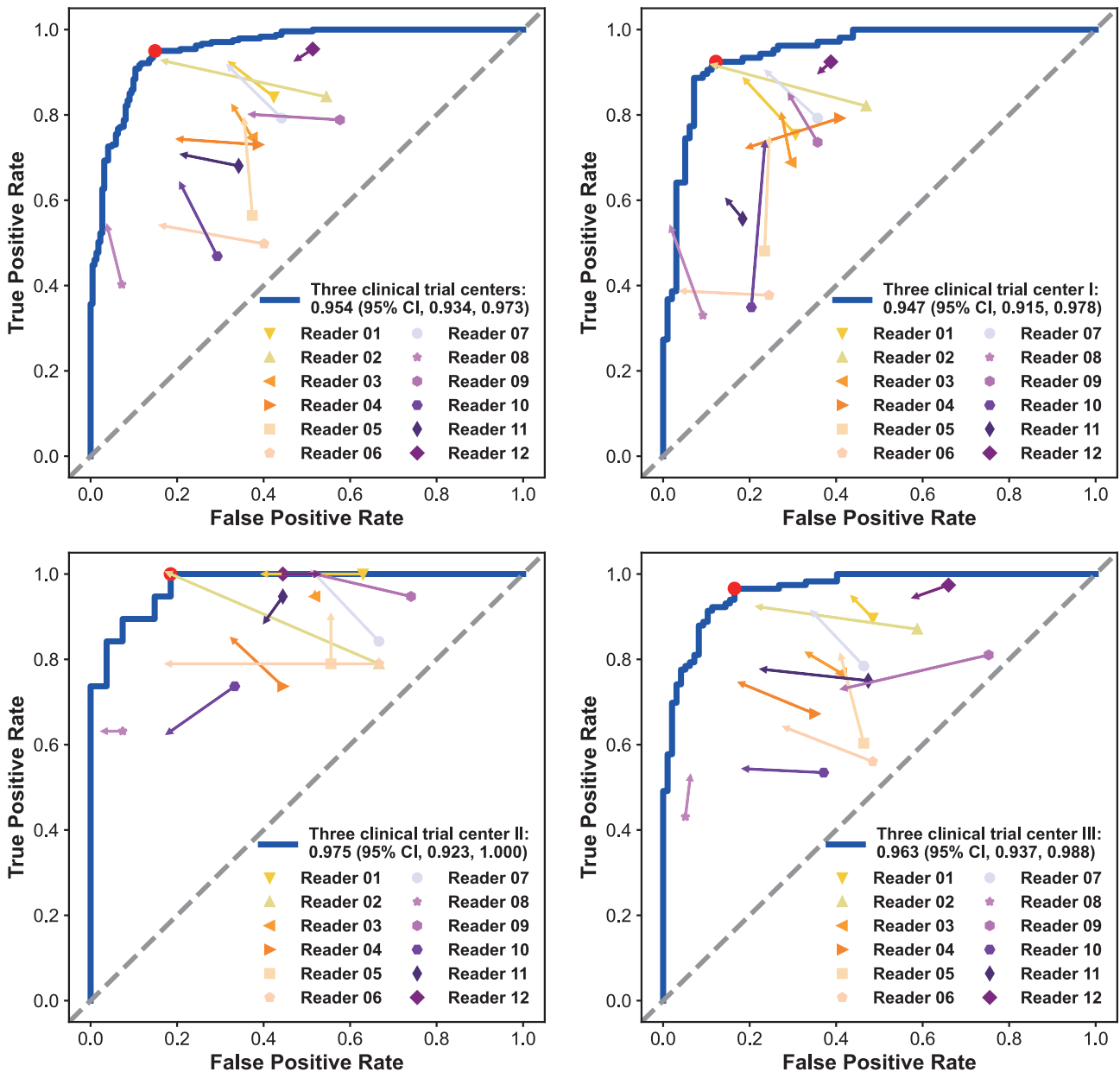

**Figure 3. ROC curves of DeepFAN and performance of readers with and without DeepFAN assistance.**
The four subplots represent the diagnostic performance of DeepFAN and the 12 readers on both the overall combined dataset from the three clinical trial centers and each center's individual dataset, respectively. The ROC curves represent the performance of DeepFAN and the red dots on the curves indicate the operating points corresponding to the binary cutoffs. The point at the base of each arrow represents the performance of each reader without DeepFAN assistance, while the arrow indicates the change in the reader's performance with DeepFAN assistance. The values in the legend are AUCs with 95%CI in parentheses.
*Abbreviations:* ROC = receiver operating characteristic, AUC = area under curve, CI = confidence interval.

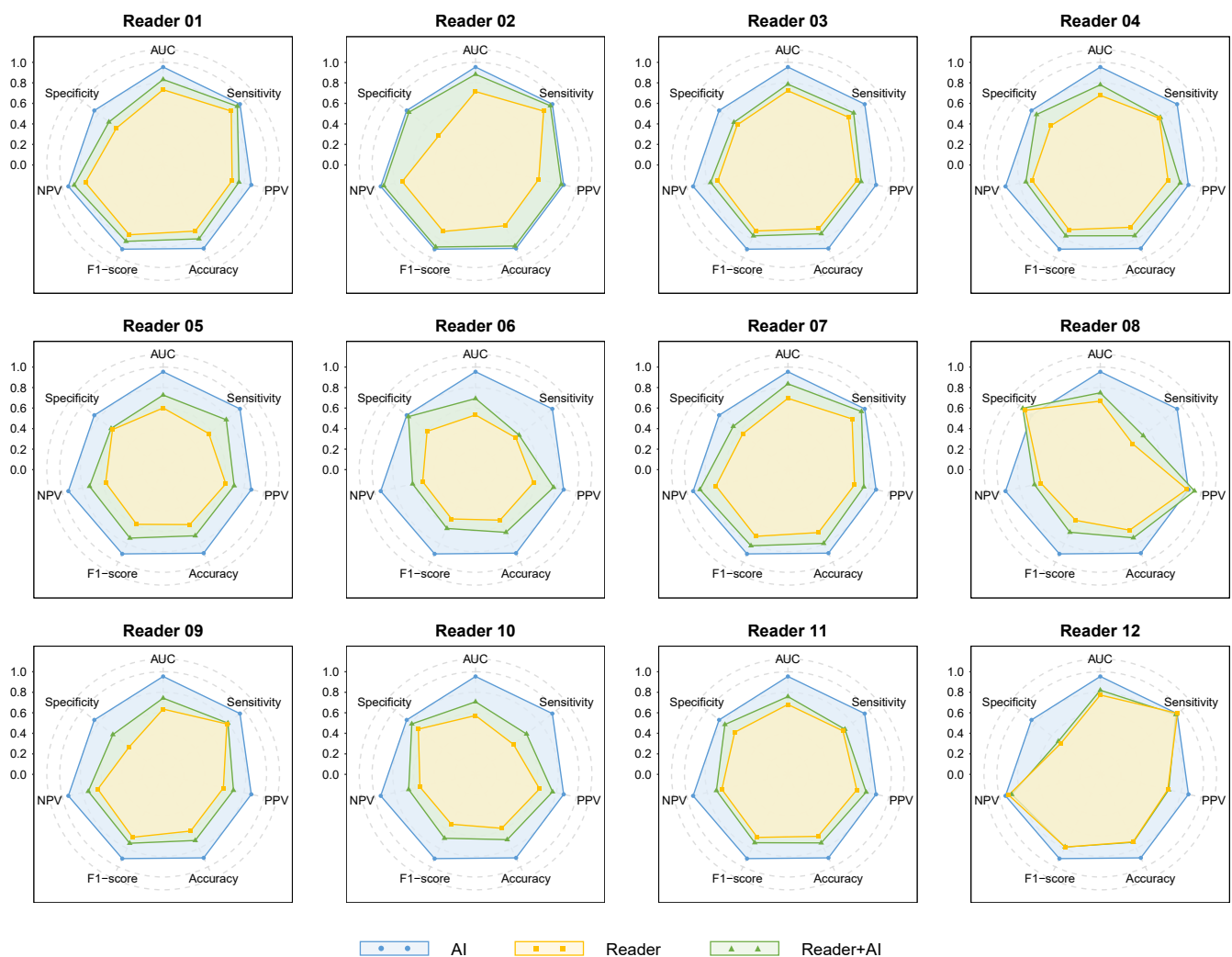

**Figure 4. Radar maps of the seven-performance metrics (AUC, sensitivity, specificity, accuracy, PPV, NPV, and F1-score) for DeepFAN and readers with and without DeepFAN assistance.**
The blue regions denote the performance indices of the AI model, the yellow regions represent reader performance without AI assistance, and the red regions illustrate reader performance aided by DeepFAN.
*Abbreviations:* AUC = area under curve, PPV = positive predictive value, NPV = negative predictive value.

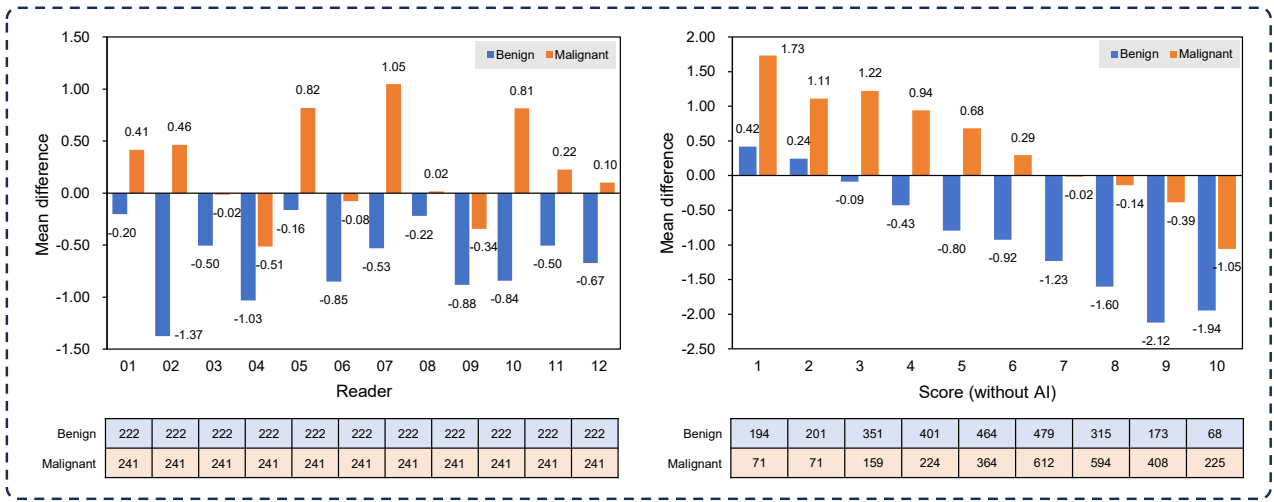

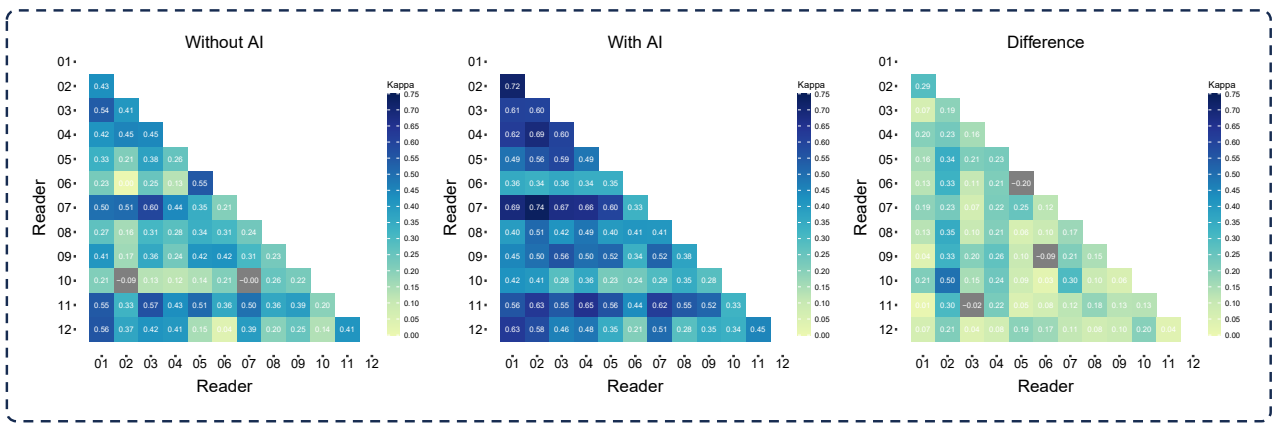

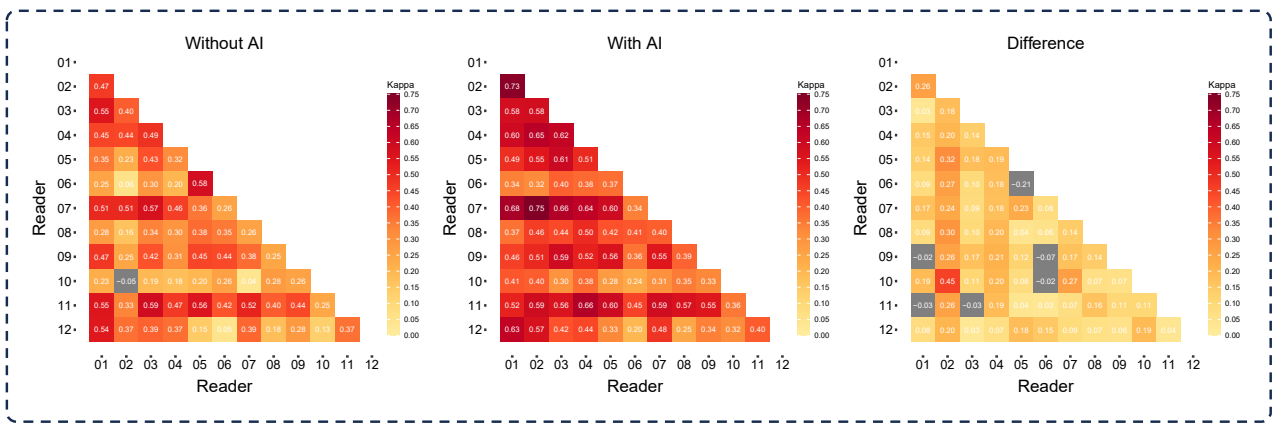

**Figure 5. Diagnostic change with AI-assisted reading and diagnostic agreement for the twelve readers.**
(a) The horizontal axis indicates different readers (left panel) and diagnostic score of unassisted reading (right panel), while the vertical axis indicates the mean differences of diagnostic scores calculated by first subtracting diagnostic scores assigned without DeepFAN assistance from those assigned with DeepFAN assistance and then averaging the score differences. The number of actual data points was illustrated in the tables below. (b, c) The numbers shown in the figure represent kappa values. These kappa values are interpreted as follows: <0.2, poor consistency; 0.21-0.4, fair consistency; 0.41-0.6, moderate consistency; 0.61-0.8, substantial consistency; 0.81-1.0, almost perfect consistency.
*Abbreviations:* AI = artificial intelligence (DeepFAN).

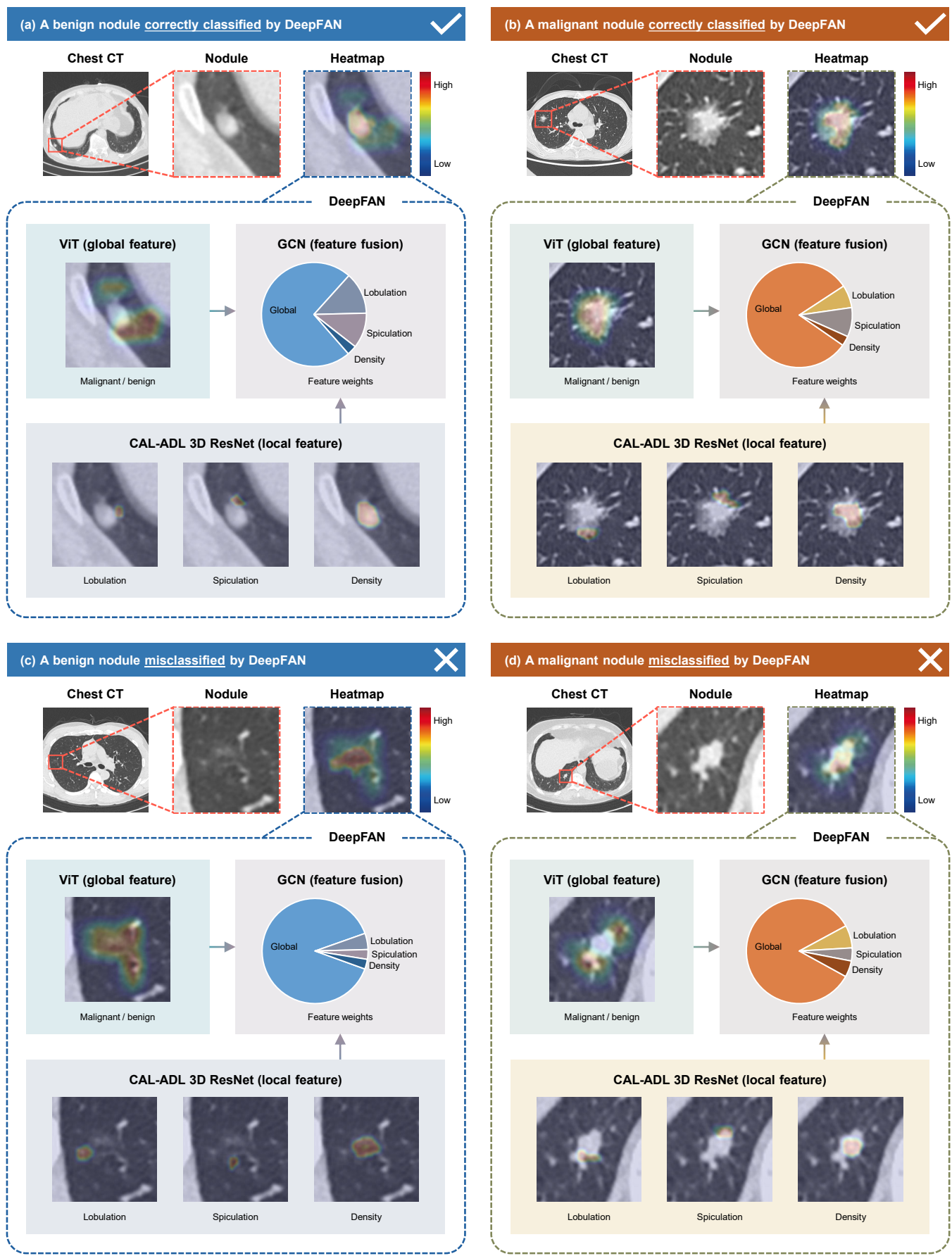

**Figure 6. Visualization of the mechanism of DeepFAN in differentiating malignant pulmonary nodules from benign ones in clinical trials.**

(a) A benign nodule (hamartoma) correctly classified by DeepFAN, with a smooth margin and fat density. (b) A

malignant nodule (invasive adenocarcinoma) correctly classified by DeepFAN, characterized by irregular shape, heterogeneous density, lobulation, and spiculation. (c) A benign nodule (epithelial hyperplasia) misclassified by DeepFAN, showing dispersed morphology, heterogeneous ground-glass density, and adjacent vascular branches. (d) A malignant nodule (invasive adenocarcinoma) misclassified by DeepFAN, with irregular shape, lobulation, spiculation, and solid density, making it difficult to differentiate from granulomatous inflammation.

In each subplot, the top row shows the original chest CT, magnified nodule images, and corresponding class activation maps generated by overlaying colored attention maps on the original image. Darker red regions signify heightened attention from DeepFAN, while darker blue regions denote reduced attention. The dashed box below shows heatmaps of each module in DeepFAN. ViT captures global features of the nodules and their surroundings, CAL-ADL 3D ResNet extracts representative local features (lobulation, spiculation, and density), and GCN is used for feature fusion. Different features (nodes) have different contributions (weights) to the final prediction result, which is calculated using the gradient, the feature value and edge weights from the first input layer of the GCN model. Subsequently, averaging and normalization are performed, yielding 12 values for 12 nodes. Notably, the global weight is determined by summing the feature weights of the 9 nodes associated with the ViT, which is the largest compared to the other nodes.

*Abbreviations:* ViT = vision transformer, CAL = counterfactual attention learning, ADL = attention dropout layer, 3D = three-dimensional, ResNet = residual network, GCN = graph convolution network.

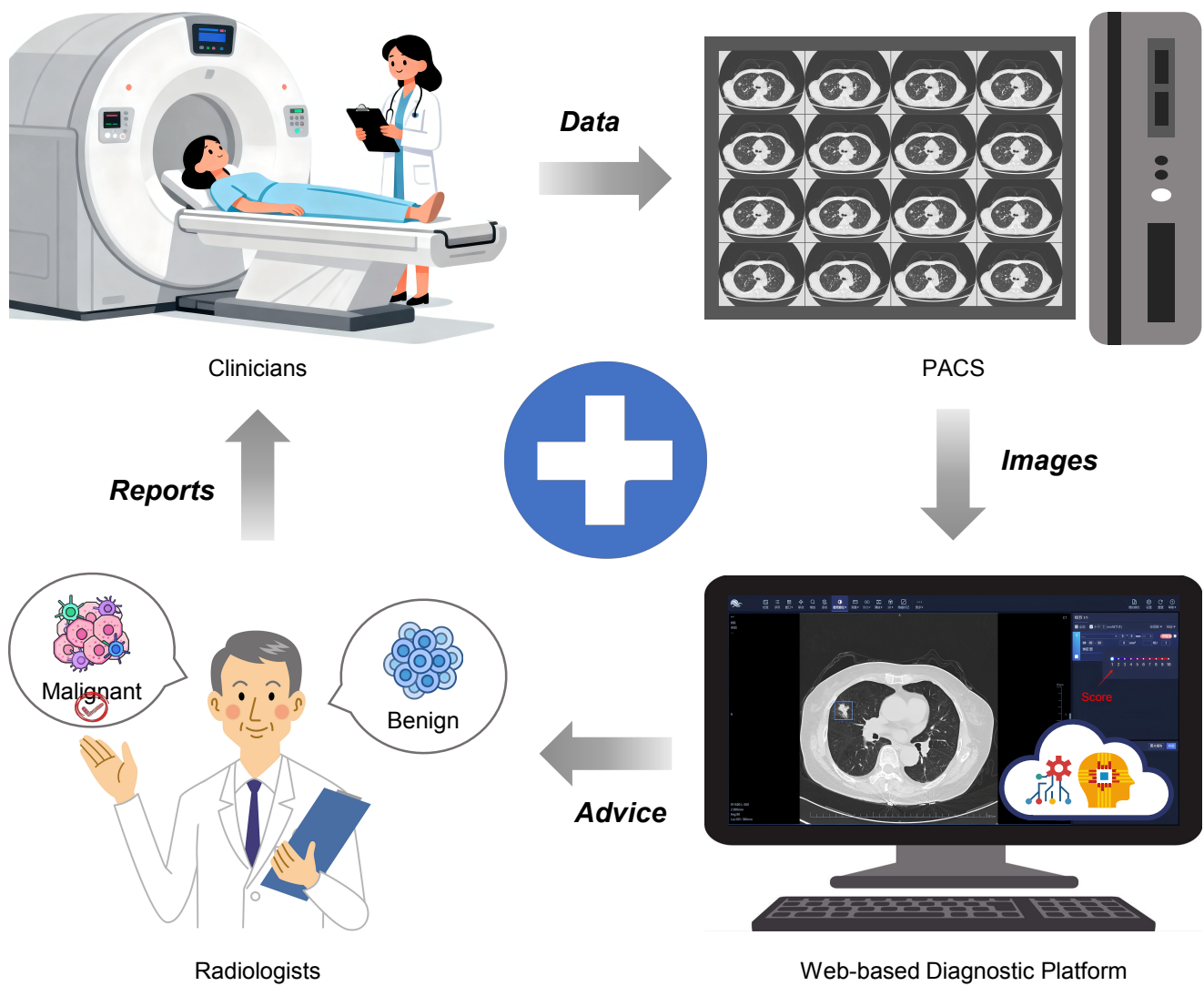

**Figure 7. Web-based AI platform for assisting the diagnosis of pulmonary nodules.**

When patients complete a CT examination under the arrangement of clinicians, their information, including chest CT images will be automatically uploaded to the PACS. Radiologists can upload corresponding CT images from PACS to the web-based AI platform. Based on the DeepFAN model, the platform will provide diagnostic advice to radiologists, including the benign and malignant classification of pulmonary nodules and the characteristics of nodules such as lobulation and spiculation. Finally, radiologists make the conclusion about the nature of pulmonary nodules after referring to this information and provide the reports to patients and clinicians.

*Abbreviations:* AI = artificial intelligence, CT = computed tomography, PACS = picture archiving and communication system.

# Supplementary Information (DeepFAN Clinical Trial Study)

# Contents

**Supplementary Table 1. Baseline characteristics of patients and pulmonary nodules in training, validation, internal test and NLST test sets**

| Variable name | Training set | Validation set | Internal test set | NLST test set |
|---|---|---|---|---|
| **Patient characteristics** | | | | |
| Total No. of patients | 5636 | 831 | 1705 | 7934 |
| Age (year)* | 57 ± 11 | 57 ± 12 | 57 ± 11 | 62 ± 5 |
| Sex | | | | |
|   Male | 2314 (41.06) | 333 (40.07) | 681 (39.94) | 4756 (59.94) |
|   Female | 3319 (58.89) | 498 (59.93) | 1023 (60.00) | 3178 (40.06) |
|   Unknown | 3 (0.05) | 0 (0.00) | 1 (0.06) | 0 (0.00) |
| Nodule type | | | | |
|   Single | 5041 (89.44) | 744 (89.53) | 1526 (89.50) | 6446 (81.25) |
|   Multiple[#] | 595 (10.56) | 87 (10.47) | 179 (10.50) | 1488 (18.75) |
| **Nodule characteristics** | | | | |
| Total no. of nodules | 7873 | 1216 | 2349 | 17892 |
| Nodule diameter (mm)* | 13.78 ± 6.75 | 13.55 ± 6.60 | 13.72 ± 6.70 | 6.58 ± 3.98 |
| Nodule density | | | | |
|   SN | 2942 (37.37) | 443 (36.43) | 937 (39.89) | 13256 (74.09) |
|   PSN | 2554 (32.44) | 358 (29.44) | 712 (30.31) | 684 (3.82) |
|   GGN | 2377 (30.19) | 415 (34.13) | 700 (29.80) | 3952 (22.09) |
| Nodule location | | | | |
|   RUL | 2805 (35.63) | 408 (33.55) | 810 (34.48) | 4542 (25.39) |
|   RML | 431 (5.47) | 74 (6.09) | 122 (5.19) | 1787 (9.99) |
|   RLL | 1467 (18.63) | 212 (17.43) | 417 (17.75) | 3688 (20.61) |
|   LUL | 1957 (24.86) | 328 (26.97) | 589 (25.07) | 3971 (22.19) |
|   LLL | 1173 (14.90) | 187 (15.38) | 397 (16.90) | 3587 (20.05) |
|   Peri-fissure | 40 (0.51) | 7 (0.58) | 14 (0.60) | 317 (1.77) |
| Pathology | | | | |
|   Benign | 1718 (21.82) | 254 (20.89) | 600 (25.54) | 16821 (94.01) |
|   Malignant | 6155 (78.18) | 962 (79.11) | 1749 (74.46) | 1071 (5.99) |

Unless otherwise indicated, numbers here are counts or percentages (in parentheses).

* Data are mean ± standard deviation. [#]Number of nodules per case ranges from two to eleven.

*Abbreviations:* NLST = national lung screening trial, SN = solid nodule, PSN = part-solid nodule, GGN = ground-glass nodule, RUL = right upper lobe, RML = right middle lobe, RLL = right lower lobe, LUL = left upper lobe, LLL = left lower lobe.

**Supplementary Table 2. Parameters of CT image series in training, validation, internal test and NLST test sets**

| Parameter | Training set | Validation set | Internal test set | NLST test set |
|---|---|---|---|---|
| Total no. of image series* | 6931 | 1053 | 2073 | 16016 |
| CT manufacture | | | | |
|   Canon | 38 (0.55) | 4 (0.38) | 1 (0.05) | 0 (0.00) |
|   GE | 1029 (14.85) | 137 (13.01) | 262 (12.64) | 2161 (13.49) |
|   NMS | 1 (0.01) | 1 (0.09) | 0 (0.00) | 0 (0.00) |
|   Philips | 1980 (28.57) | 280 (26.59) | 591 (28.51) | 474 (2.96) |
|   SIEMENS | 3379 (48.75) | 544 (51.66) | 1030 (49.69) | 11724 (73.20) |
|   SinoVision | 101 (1.46) | 16 (1.52) | 26 (1.25) | 0 (0.00) |
|   TOSHIBA | 244 (3.52) | 34 (3.23) | 88 (4.25) | 1657 (10.35) |
|   UIH | 154 (2.22) | 37 (3.51) | 70 (3.38) | 0 (0.00) |
|   Unknown | 5 (0.07) | 0 (0.00) | 5 (0.24) | 0 (0.00) |
| Slice thickness (mm) | | | | |
|   ≥0.5 to <1 | 1164 (16.79) | 154 (14.62) | 298 (14.38) | 12 (0.07) |
|   1 | 5222 (75.34) | 816 (77.49) | 1630 (78.63) | 369 (2.30) |
|   >1 to ≤2 | 545 (7.86) | 83 (7.88) | 145 (6.99) | 15635 (97.62) |
| Reconstruction kernel | | | | |
|   Lung | 3590 (51.80) | 539 (51.19) | 1015 (48.96) | 712 (4.45) |
|   Mediastinum | 1310 (18.90) | 203 (19.28) | 429 (20.69) | 9136 (57.04) |
|   Bone | 1835 (26.48) | 287 (27.26) | 573 (27.64) | 5204 (32.49) |
|   Other | 196 (2.83) | 24 (2.28) | 56 (2.70) | 964 (6.02) |
| Image matrix | | | | |
|   512×512 | 5250 (75.75) | 803 (76.26) | 1544 (74.48) | 16016 (100.00) |
|   768×768 | 13 (0.19) | 2 (0.19) | 12 (0.58) | 0 (0.00) |
|   1024×1024 | 1668 (24.07) | 248 (23.55) | 517 (24.94) | 0 (0.00) |
| Tube voltage (kVp) | | | | |
|   80 | 0 (0.00) | 2 (0.19) | 2 (0.10) | 1 (0.01) |
|   90 | 15 (0.22) | 3 (0.28) | 2 (0.10) | 0 (0.00) |
|   100 | 146 (2.11) | 32 (3.04) | 67 (3.23) | 2 (0.01) |
|   110 | 232 (3.35) | 31 (2.94) | 65 (3.14) | 2 (0.01) |
|   120 | 5876 (84.78) | 877 (83.29) | 1751 (84.47) | 15598 (97.39) |
|   130 | 540 (7.79) | 85 (8.07) | 137 (6.61) | 15 (0.09) |
|   140 | 96 (1.39) | 17 (1.61) | 40 (1.93) | 398 (2.49) |
|   150 | 21 (0.30) | 6 (0.57) | 4 (0.19) | 0 (0.00) |
|   Unknown | 5 (0.07) | 0 (0.00) | 5 (0.24) | 0 (0.00) |

Numbers here are counts or percentages (in parentheses).

* A patient may have multiple image series with different slice thicknesses or reconstruction kernels.

*Abbreviations:* NLST = national lung screening trial, Canon = Canon Medical Systems, GE = General Electric Healthcare, NMS = Neusoft Medical Systems, Philips = Philips Healthcare, Siemens = Siemens Healthineers, SinoVision = SinoVision Technology, UIH = United Imaging Healthcare.

**Supplementary Table 3. Diagnostic performance of DeepFAN on the internal test set, clinical trial and NLST test sets**

| Dataset | | AUC | Sensitivity | Specificity | Accuracy | PPV | NPV | F1-score |
|---|---|---|---|---|---|---|---|---|
| Internal test set | | 0.939 (0.930, 0.948) | 0.953 (0.943, 0.962) | 0.733 (0.699, 0.768) | 0.897 (0.884, 0.909) | 0.912 (0.899, 0.926) | 0.841 (0.813, 0.871) | 0.932 (0.923, 0.940) |
| Clinical trial dataset* | | 0.954 (0.934, 0.973) | 0.950 (0.923, 0.978) | 0.851 (0.805, 0.898) | 0.903 (0.876, 0.930) | 0.874 (0.834, 0.913) | 0.940 (0.905, 0.970) | 0.911 (0.883, 0.935) |
| | Center I | 0.947 (0.915, 0.978) | 0.925 (0.874, 0.972) | 0.878 (0.813, 0.938) | 0.902 (0.863, 0.941) | 0.891 (0.833, 0.944) | 0.915 (0.856, 0.969) | 0.908 (0.860, 0.943) |
| | Center II | 0.975 (0.923, 1.000) | 1.000 (1.000, 1.000) | 0.815 (0.640, 0.957) | 0.891 (0.783, 0.978) | 0.792 (0.600, 0.955) | 1.000 (1.000, 1.000) | 0.884 (0.735, 0.962) |
| | Center III | 0.963 (0.937, 0.988) | 0.966 (0.932, 0.992) | 0.835 (0.762, 0.905) | 0.906 (0.864, 0.944) | 0.875 (0.816, 0.928) | 0.953 (0.905, 0.989) | 0.918 (0.876, 0.951) |
| NLST test set | | 0.943 (0.933, 0.953) | 0.889 (0.869, 0.908) | 0.897 (0.893, 0.902) | 0.897 (0.892, 0.901) | 0.356 (0.338, 0.374) | 0.992 (0.991, 0.994) | 0.508 (0.489, 0.528) |

Numbers here are diagnostic performance measures with 95% confidence interval in parentheses.

*Clinical trial dataset contains data from all three clinical trial centers.

*Abbreviations:* AUC = area under curve, PPV = positive predictive value, NPV = negative predictive value, NLST = national lung screening trial.

**Supplementary Table 4. Diagnostic performance of DeepFAN in ablation experiments on the internal test set**

| Model* | Ablation experiments[†] | | | Diagnostic performance | | | | | | |
|---|---|---|---|---|---|---|---|---|---|---|
| | Global feature | Local feature | Feature fusion | AUC | Sensitivity | Specificity | Accuracy | PPV | NPV | F1-score |
| Model 1 | ViT | … | … | 0.879 (0.866, 0.893) | 0.881 (0.866, 0.896) | 0.650 (0.611, 0.687) | 0.822 (0.806, 0.837) | 0.880 (0.864, 0.896) | 0.652 (0.618, 0.690) | 0.881 (0.869, 0.892) |
| Model 2 | ResNet | … | … | 0.861 (0.846, 0.875) | 0.872 (0.856, 0.888) | 0.628 (0.591, 0.666) | 0.810 (0.792, 0.825) | 0.872 (0.855, 0.889) | 0.627 (0.588, 0.666) | 0.872 (0.859, 0.884) |
| Model 3 | CAL-ADL | … | … | 0.863 (0.848, 0.877) | 0.863 (0.846, 0.878) | 0.647 (0.609, 0.683) | 0.808 (0.791, 0.823) | 0.877 (0.861, 0.892) | 0.619 (0.581, 0.656) | 0.870 (0.858, 0.882) |
| Model 4 | ViT | CAL-ADL | Concat | 0.904 (0.892, 0.916) | 0.923 (0.911, 0.935) | 0.668 (0.631, 0.705) | 0.858 (0.843, 0.872), | 0.890 (0.874, 0.905) | 0.748 (0.713, 0.784) | 0.906 (0.895, 0.916) |
| Model 5 | ViT | ViT | GCN | 0.927 (0.917, 0.937) | 0.952 (0.942, 0.961) | 0.668 (0.631, 0.705) | 0.880 (0.865, 0.892) | 0.893 (0.878, 0.907) | 0.827 (0.796, 0.859) | 0.922 (0.912, 0.931) |
| Model 6 | ViT | ResNet50 | GCN | 0.920 (0.909, 0.930) | 0.947 (0.937, 0.958) | 0.668 (0.632, 0.704) | 0.876 (0.862, 0.889) | 0.893 (0.877, 0.907) | 0.813 (0.781, 0.847) | 0.919 (0.909, 0.928) |
| Model 7 | CAL-ADL | CAL-ADL | GCN | 0.913 (0.902, 0.924) | 0.941 (0.930, 0.951) | 0.668 (0.631, 0.705) | 0.871 (0.857, 0.885) | 0.892 (0.877, 0.906) | 0.794 (0.762, 0.827) | 0.916 (0.905, 0.925) |
| Model 8 | ResNet50 | CAL-ADL | GCN | 0.901 (0.889, 0.913) | 0.913 (0.900, 0.926) | 0.667 (0.629, 0.703) | 0.850 (0.834, 0.863) | 0.889 (0.872, 0.903) | 0.723 (0.686, 0.760) | 0.900 (0.889, 0.910) |
| Model 9 (DeepFAN) | ViT | CAL-ADL | GCN | 0.939 (0.930, 0.948) | 0.953 (0.943, 0.962) | 0.733 (0.699, 0.768) | 0.897 (0.884, 0.909) | 0.912 (0.899, 0.926) | 0.841 (0.813, 0.871) | 0.932 (0.923, 0.940) |

Numbers here are diagnostic performance measures with 95% confidence interval in parentheses.

* The cut-off value for all models in ablation experiments is 0.50.

[†] To assess the contribution of each module within DeepFAN to its overall performance, ablation experiments were conducted on the internal test set using ViT, ResNet50, and CAL-ADL 3D ResNet as component networks. The procedure involved progressively removing, adapting, and substituting key components of DeepFAN to evaluate their individual impact on the system efficacy.

*Abbreviations:* AUC = area under curve, PPV = positive predictive value, NPV = negative predictive value, ViT = vision transformer, ResNet = residual network, CAL-ADL = three-dimensional residual network based on counterfactual attention learning and attention dropout layer, Concat = concatenate, GCN = graph convolution network.

**Supplementary Table 5. Performance comparison between DeepFAN and models reported in previous studies on pulmonary nodule diagnosis**

| Models or datasets | AUC | Sensitivity | Specificity | Accuracy | PPV | NPV | F1-score |
|---|---|---|---|---|---|---|---|
| **DeepFAN model (this study)** | | | | | | | |
| NLST dataset | 0.943 (0.933, 0.953) | 0.889 (0.869, 0.908) | 0.897 (0.893, 0.902) | 0.897 (0.892, 0.901) | 0.356 (0.338, 0.374) | 0.992 (0.991, 0.994) | 0.508 (0.489, 0.528) |
| NLST dataset (without GGN)† | 0.949 (0.939, 0.959) | 0.889 (0.871, 0.907) | 0.916 (0.911, 0.921) | 0.914 (0.931, 0.951) | 0.474 (0.453, 0.494) | 0.990 (0.988, 0.992) | 0.618 (0.597, 0.637) |
| **LCP-CNN model (Am J Respir Crit Care Med, 2020)[8]** | | | | | | | |
| NLST dataset (without GGN)‡ | 0.921 (0.912, 0.929) | 0.956 (0.942, 0.969) | 0.629 (0.621, 0.637) | 0.648* | 0.140* | 0.995* | 0.244* |
| **Mayo model (Arch Intern Med, 1997)[37]** | | | | | | | |
| NLST dataset (without GGN)† used in this study | 0.852 (0.837, 0.866) | 0.665 (0.646, 0.683) | 0.871 (0.866, 0.877) | 0.855 (0.842, 0.867) | 0.304 (0.283, 0.324) | 0.968 (0.965, 0.970) | 0.418 (0.398, 0.437) |
| NLST dataset (without GGN)‡ used in the LCP-CNN study | 0.852 (0.841, 0.862) | NA | NA | NA | NA | NA | NA |
| **Brock model (N Engl J Med, 2013)[36]** | | | | | | | |
| NLST dataset (without GGN)† used in this study | 0.856 (0.840, 0.871) | 0.865 (0.843, 0.886) | 0.646 (0.638, 0.654) | 0.661 (0.654, 0.669) | 0.153 (0.143, 0.163) | 0.985 (0.982, 0.987) | 0.260 (0.245, 0.274) |
| NLST dataset (without GGN)‡ used in the LCP-CNN study | 0.856 (0.843, 0.868) | 0.865 (0.841, 0.886) | 0.665 (0.658, 0.672) | NA | NA | NA | NA |
| **DL model (Radiology, 2021)[17]** | | | | | | | |
| NLST dataset‖ | 0.910 (0.900, 0.920) | 0.710 (0.700, 0.720) | 0.900 (0.890, 0.910) | 0.885* | 0.374* | 0.974* | 0.489* |
| **DCNN model (Nat Med, 2024)[22]** | | | | | | | |
| MCC dataset[a] | 0.918 (0.918, 0.919) | 0.851 (0.850, 0.853) | 0.828 (0.828, 0.829) | 0.830* | 0.473* | 0.967* | 0.607* |
| MSC dataset[b] | 0.927 (0.926, 0.928) | 0.856 (0.851, 0.861) | 0.877 (0.876, 0.877) | 0.874* | 0.434* | 0.981* | 0.574* |

The performance of DeepFAN was compared with previous methods for pulmonary nodule diagnosis by gathering performance metrics reported in published papers. It is important to note that in this comparison, the NLST datasets used by different models were not identical. Testing all considered models on the same dataset was infeasible due to the unavailability of trained models or training data from those studies.

\* These performance measures were not directly available from previous studies but were derived from other metrics reported in the literature.

† To facilitate a more direct comparison with the LCP-CNN study (Am J Respir Crit Care Med, 2020), the GGNs were excluded from the NLST dataset used in this study, creating a new test set. The new NLST dataset included 12978 benign nodules and 957 malignant nodules, with nodule sizes ranging from 5 to 30 mm.

‡ The NLST dataset used in the LCP-CNN study (Am J Respir Crit Care Med, 2020) included 14761 benign nodules (5972 patients) and 932 malignant nodules (575 patients). These nodules did not include GGNs and ranged in size from 5 to 30 mm.

∥ The NLST dataset used in the DL model study (Radiology, 2021) included 14828 benign nodules and 1249 malignant nodules, with nodule sizes > 4mm.

a An internal testing dataset used in the DCNN study (Nat Med, 2024). The dataset was obtained from the medical checkup cohort (MCC) at the health management center in West China Hospital of Sichuan University and included 1142 benign nodules and 209 malignant nodules.

b An external testing dataset in the DCNN study (Nat Med, 2024). The dataset was obtained from a mobile screening cohort (MSC) across multiple communities in Western China and included 1812 benign nodules and 139 malignant nodules.

*Abbreviations:* AUC = area under curve, PPV = positive predictive value, NPV = negative predictive value, NLST = national lung screening trial, GGN = ground-glass nodule, LCP-CNN = lung cancer prediction convolutional neural network, DL = deep learning, DCNN = deep convolutional neural network, NA = not available.

**Supplementary Table 6. Baseline characteristics of benign and malignant patients and nodules in datasets from three clinical trial centers**

| Variable name | Benign | Malignant | *P* value |
|---|---|---|---|
| **Patient characteristics** | | | |
| Total no. of patients | 197 | 203 | |
| Age (year)* | 58 ± 10 | 61 ± 10 | 0.005 |
| Sex | | | 0.563 |
|   Male | 112 (57) | 89 (44) | |
|   Female | 85 (43) | 114 (56) | |
| Nodule type | | | 0.041 |
|   Single | 176 (89) | 176 (87) | |
|   Multiple# | 21 (11) | 27 (13) | |
| **Nodule characteristics** | | | |
| Total no. of nodules | 222 | 241 | |
| Nodule diameter (mm)* | 11.55 ± 5.44 | 15.19 ± 5.87 | <.001 |
| Nodule density | | | <.001 |
|   SN | 133 (60) | 59 (24) | |
|   PSN | 36 (16) | 132 (55) | |
|   GGN | 53 (24) | 50 (21) | |
| Nodule location | | | 0.427 |
|   RUL | 58 (26) | 67 (28) | |
|   RML | 17 (8) | 19 (8) | |
|   RLL | 58 (26) | 58 (24) | |
|   LUL | 41 (18) | 58 (24) | |
|   LLL | 48 (22) | 39 (16) | |
| Spiculation | | | <.001 |
|   No | 139 (63) | 69 (29) | |
|   Yes | 83 (37) | 172 (71) | |
| Lobulation | | | <.001 |
|   No | 56 (25) | 14 (6) | |
|   Yes | 166 (75) | 227 (94) | |

Unless otherwise indicated, numbers here are counts or percentages (in parentheses).

* Data are mean ± standard deviation. #Number of nodules per case ranges from two to seven.

*Abbreviations:* SN = solid nodule, PSN = part-solid nodule, GGN = ground-glass nodule, RUL = right upper lobe, RML = right middle lobe, RLL = right lower lobe, LUL = left upper lobe, LLL = left lower lobe.

**Supplementary Table 7. Characteristics of chest CT images in datasets from three clinical trial centers**

| Parameter | Clinical trial center I (n=166) | Clinical trial center II (n=46) | Clinical trial center III (n=188) |
|---|---|---|---|
| CT manufacture | | | |
| Siemens | 128 (77) | 3 (7) | 16 (9) |
| GE | 37 (22) | 12 (26) | 168 (89) |
| UIH | 0 (0) | 31 (88) | 4 (12) |
| Others | 1 (100) | 0 (0) | 0 (0) |
| Detectors | | | |
| 2*96 | 128 (77) | 0 (0) | 0 (0) |
| 64 | 0 (0) | 19 (41) | 47 (25) |
| 80 | 1 (1) | 15 (33) | 0 (0) |
| 128 | 0 (0) | 9 (20) | 0 (0) |
| 256 | 37 (22) | 3 (7) | 137 (73) |
| 16 | 0 (0) | 0 (0) | 3 (2) |
| 40 | 0 (0) | 0 (0) | 1 (1) |
| Slice thickness (mm) | | | |
| ≥0.625 to <1 | 0 (0) | 10 (22) | 183 (97) |
| 1 | 129 (78) | 33 (72) | 2 (1) |
| >1 to ≤1.25 | 37 (22) | 3 (7) | 3 (2) |
| Reconstruction kernel | | | |
| Br40d-3 | 128 (77) | 0 (0) | 0 (0) |
| STANDARD | 37 (22) | 11 (24) | 32 (17) |
| B-SOFT-B | 1 (1) | 5 (11) | 3 (2) |
| B-SHARP-C | 0 (0) | 22 (48) | 1 (1) |
| I70f.3 | 0 (0) | 2 (4) | 0 (0) |
| LUNG | 0 (0) | 1 (2) | 136 (72) |
| B-SOFT-C | 0 (0) | 4 (9) | 0 (0) |
| 13'f'3 | 0 (0) | 1 (2) | 0 (0) |
| B31s | 0 (0) | 0 (0) | 16 (9) |

Numbers here are counts or percentages (in parentheses). n in the parentheses refers to the total number of patients at each center.

*Abbreviations:* Siemens = Siemens Healthineers, GE = General Electric Healthcare, UIH = United Imaging Healthcare.

**Supplementary Table 8. Baseline characteristics of twelve readers**

| Number | Gender | Age (years old) | Center name | Internship# (months) | Working experience | Education | Major | Group* |
|---|---|---|---|---|---|---|---|---|
| Reader 01 | Female | 31 | Clinical trial center I | 12 | 2 years | MD | Medical imaging | B |
| Reader 02 | Female | 29 | Clinical trial center I | 12 | 1 year | MD | Medical imaging | B |
| Reader 03 | Female | 29 | Clinical trial center I | 6 | 1 year | MD | Medical imaging | A |
| Reader 04 | Female | 31 | Clinical trial center II | 18 | 5 years | MM | Medical imaging | A |
| Reader 05 | Male | 28 | Clinical trial center II | 24 | 2 years | BM | Medical imaging | A |
| Reader 06 | Male | 30 | Clinical trial center II | 12 | 3 years | MM | Medical imaging | B |
| Reader 07 | Male | 31 | Clinical trial center II | 12 | 4 years | MM | Medical imaging | B |
| Reader 08 | Female | 31 | Clinical trial center III | 26 | 1 year | MM | Clinical medicine | A |
| Reader 09 | Male | 29 | Clinical trial center III | 12 | 2 years | MM | Medical imaging | A |
| Reader 10 | Female | 30 | Clinical trial center III | 12 | 4 years | BM | Medical imaging | B |
| Reader 11 | Female | 30 | Clinical trial center III | 9 | 4 years | BM | Medical imaging | B |
| Reader 12 | Female | 30 | Clinical trial center III | 12 | 5 years | BM | Medical imaging | A |

Clinical trial centers I, II, and III represent Peking University People's Hospital, Wuhan Third Hospital, and Huangshi Central Hospital, respectively. #Internship refers to any clinical practice experience prior to China's standardized residency training and is not limited to the radiology department; it may also include rotations in internal medicine, general surgery, and other departments. *Group denotes the randomly assigned group number in the multireader multicase study design (detailed under the subtitle of **Clinical trial design** in **Methods** section). More details regarding the characteristics of the twelve readers and the questionnaires related to these responses are illustrated in **Supplementary Table 14** and **reader questionnaire** in the Supplementary Information.

*Abbreviations*: BM = bachelor of medicine, MM = master of medicine, MD = doctor of medicine

**Supplementary Table 9. Comparison of diagnostic performance of the twelve readers in the clinical trial**

| Reader | | AUC | Accuracy | Sensitivity | Specificity | PPV | NPV | F1-score |
|---|---|---|---|---|---|---|---|---|
| Twelve readers | Without DeepFAN | 0.667 (0.616, 0.719) | 0.651 (0.638, 0.663) | 0.693 (0.676, 0.709) | 0.605 (0.586, 0.623) | 0.655 (0.638, 0.673) | 0.644 (0.625, 0.663) | 0.674 (0.659, 0.689) |
| | With DeepFAN | 0.776 (0.733, 0.819) | 0.751 (0.739, 0.762) | 0.769 (0.754, 0.784) | 0.731 (0.714, 0.747) | 0.756 (0.740, 0.772) | 0.744 (0.728, 0.762) | 0.762 (0.750, 0.775) |
| | Difference | 0.109 (0.083, 0.135) | 0.100 (0.089, 0.111) | 0.076 (0.061, 0.092) | 0.126 (0.109, 0.143) | 0.101 (0.089, 0.112) | 0.100 (0.087, 0.113) | 0.089 (0.078, 0.100) |
| | P-value | <0.001 | <0.001 | <0.001 | <0.001 | <0.001 | <0.001 | <0.001 |
| Reader 01 | Without DeepFAN | 0.733 (0.688, 0.778) | 0.715 (0.672, 0.756) | 0.842 (0.797, 0.887) | 0.577 (0.511, 0.641) | 0.684 (0.632, 0.734) | 0.771 (0.707, 0.831) | 0.755 (0.712, 0.793) |
| | With DeepFAN | 0.833 (0.796, 0.870) | 0.799 (0.762, 0.834) | 0.917 (0.878, 0.950) | 0.671 (0.607, 0.730) | 0.752 (0.701, 0.799) | 0.882 (0.825, 0.926) | 0.826 (0.791, 0.862) |
| | Difference | 0.100 (0.059, 0.141) | 0.084 (0.045, 0.125) | 0.075 (0.032, 0.122) | 0.095 (0.028, 0.159) | 0.068 (0.032, 0.108) | 0.111 (0.055, 0.171) | 0.072 (0.040, 0.106) |
| | P-value | <0.001 | <0.001 | 0.005 | 0.006 | <0.001 | <0.001 | <0.001 |
| Reader 02 | Without DeepFAN | 0.715 (0.669, 0.761) | 0.657 (0.616, 0.700) | 0.842 (0.795, 0.888) | 0.455 (0.390, 0.522) | 0.627 (0.579, 0.678) | 0.727 (0.653, 0.805) | 0.719 (0.673, 0.757) |
| | With DeepFAN | 0.883 (0.852, 0.914) | 0.877 (0.844, 0.907) | 0.925 (0.892, 0.957) | 0.824 (0.778, 0.870) | 0.851 (0.807, 0.890) | 0.910 (0.867, 0.947) | 0.887 (0.857, 0.915) |
| | Difference | 0.168 (0.126, 0.210) | 0.220 (0.177, 0.263) | 0.083 (0.041, 0.127) | 0.369 (0.304, 0.436) | 0.225 (0.180, 0.269) | 0.184 (0.117, 0.250) | 0.168 (0.131, 0.204) |
| | P-value | <0.001 | <0.001 | 0.001 | <0.001 | <0.001 | <0.001 | <0.001 |
| Reader 03 | Without DeepFAN | 0.723 (0.677, 0.769) | 0.689 (0.648, 0.730) | 0.747 (0.693, 0.799) | 0.626 (0.564, 0.689) | 0.684 (0.629, 0.737) | 0.695 (0.626, 0.755) | 0.714 (0.668, 0.760) |
| | With DeepFAN | 0.787 (0.746, 0.828) | 0.743 (0.702, 0.784) | 0.813 (0.761, 0.863) | 0.667 (0.604, 0.728) | 0.726 (0.675, 0.779) | 0.767 (0.705, 0.821) | 0.767 (0.724, 0.804) |
| | Difference | 0.064 (0.028, 0.100) | 0.054 (0.017, 0.093) | 0.066 (0.020, 0.116) | 0.041 (-0.017, 0.097) | 0.042 (0.006, 0.081) | 0.072 (0.027, 0.120) | 0.053 (0.020, 0.088) |
| | P-value | <0.001 | 0.006 | 0.015 | 0.2 | 0.02 | 0.004 | 0.002 |
| Reader 04 | Without DeepFAN | 0.678 (0.630, 0.726) | 0.674 (0.633, 0.717) | 0.730 (0.675, 0.784) | 0.613 (0.551, 0.676) | 0.672 (0.618, 0.727) | 0.677 (0.609, 0.739) | 0.700 (0.653, 0.749) |
| | With DeepFAN | 0.781 (0.739, 0.823) | 0.765 (0.724, 0.799) | 0.743 (0.689, 0.793) | 0.788 (0.728, 0.839) | 0.792 (0.738, 0.842) | 0.738 (0.680, 0.790) | 0.767 (0.720, 0.805) |
| | Difference | 0.103 (0.063, 0.144) | 0.091 (0.054, 0.127) | 0.012 (-0.034, 0.057) | 0.176 (0.120, 0.236) | 0.120 (0.079, 0.162) | 0.062 (0.021, 0.101) | 0.067 (0.033, 0.101) |
| | P-value | <0.001 | <0.001 | 0.728 | <0.001 | <0.001 | 0.006 | <0.001 |
| Reader 05 | Without DeepFAN | 0.601 (0.550, 0.652) | 0.594 (0.549, 0.639) | 0.564 (0.496, 0.627) | 0.626 (0.561, 0.687) | 0.621 (0.555, 0.683) | 0.570 (0.504, 0.627) | 0.591 (0.536, 0.645) |
| | With DeepFAN | 0.727 (0.681, 0.773) | 0.715 (0.674, 0.756) | 0.780 (0.725, 0.832) | 0.644 (0.582, 0.706) | 0.704 (0.650, 0.760) | 0.730 (0.667, 0.789) | 0.740 (0.694, 0.781) |
| | Difference | 0.126 (0.085, 0.167) | 0.121 (0.080, 0.160) | 0.216 (0.161, 0.272) | 0.018 (-0.038, 0.077) | 0.083 (0.041, 0.125) | 0.160 (0.115, 0.204) | 0.149 (0.107, 0.191) |
| | P-value | <0.001 | <0.001 | <0.001 | 0.643 | <0.001 | 0.006 | <0.001 |
| Reader 06 | Without DeepFAN | 0.533 (0.480, 0.586) | 0.546 (0.501, 0.592) | 0.498 (0.436, 0.559) | 0.599 (0.535, 0.661) | 0.574 (0.507, 0.643) | 0.524 (0.460, 0.581) | 0.533 (0.478, 0.590) |
| | With DeepFAN | 0.693 (0.645, 0.741) | 0.678 (0.635, 0.719) | 0.539 (0.474, 0.602) | 0.829 (0.778, 0.876) | 0.774 (0.709, 0.834) | 0.624 (0.567, 0.677) | 0.636 (0.578, 0.689) |
| | Difference | 0.160 (0.108, 0.213) | 0.132 (0.080, 0.181) | 0.041 (-0.038, 0.119) | 0.230 (0.171, 0.291) | 0.200 (0.143, 0.258) | 0.100 (0.053, 0.148) | 0.102 (0.034, 0.166) |

| Reader | Measure | | | | | | | |
|---|---|---|---|---|---|---|---|---|
| | P-value | <0.001 | <0.001 | 0.343 | <0.001 | <0.001 | <0.001 | <0.001 |
| Reader 07 | Without DeepFAN | 0.695 (0.648, 0.742) | 0.680 (0.635, 0.721) | 0.793 (0.743, 0.841) | 0.559 (0.491, 0.623) | 0.661 (0.608, 0.712) | 0.713 (0.642, 0.778) | 0.721 (0.673, 0.766) |
| | With DeepFAN | 0.836 (0.799, 0.873) | 0.797 (0.760, 0.832) | 0.909 (0.871, 0.942) | 0.676 (0.612, 0.734) | 0.753 (0.703, 0.800) | 0.872 (0.819, 0.918) | 0.823 (0.785, 0.857) |
| | Difference | 0.141 (0.105, 0.177) | 0.117 (0.084, 0.156) | 0.116 (0.074, 0.160) | 0.117 (0.064, 0.169) | 0.092 (0.061, 0.128) | 0.159 (0.107, 0.213) | 0.103 (0.074, 0.136) |
| | P-value | <0.001 | <0.001 | <0.001 | <0.001 | <0.001 | <0.001 | <0.001 |
| Reader 08 | Without DeepFAN | 0.668 (0.619, 0.717) | 0.654 (0.611, 0.695) | 0.402 (0.339, 0.466) | 0.928 (0.892, 0.958) | 0.858 (0.791, 0.919) | 0.589 (0.532, 0.636) | 0.548 (0.486, 0.608) |
| | With DeepFAN | 0.749 (0.705, 0.793) | 0.737 (0.695, 0.775) | 0.531 (0.467, 0.591) | 0.959 (0.932, 0.985) | 0.934 (0.892, 0.975) | 0.653 (0.599, 0.701) | 0.677 (0.614, 0.731) |
| | Difference | 0.080 (0.046, 0.115) | 0.082 (0.050, 0.115) | 0.129 (0.071, 0.183) | 0.032 (-0.005, 0.071) | 0.076 (0.015, 0.142) | 0.065 (0.039, 0.091) | 0.129 (0.077, 0.184) |
| | P-value | 0.004 | <0.001 | <0.001 | 0.146 | 0.014 | <0.001 | <0.001 |
| Reader 09 | Without DeepFAN | 0.636 (0.586, 0.686) | 0.613 (0.570, 0.659) | 0.788 (0.735, 0.839) | 0.423 (0.364, 0.489) | 0.597 (0.548, 0.651) | 0.648 (0.568, 0.726) | 0.680 (0.632, 0.721) |
| | With DeepFAN | 0.743 (0.699, 0.787) | 0.715 (0.674, 0.756) | 0.801 (0.750, 0.850) | 0.622 (0.560, 0.685) | 0.697 (0.643, 0.750) | 0.742 (0.676, 0.801) | 0.745 (0.703, 0.786) |
| | Difference | 0.106 (0.070, 0.143) | 0.102 (0.060, 0.140) | 0.012 (-0.031, 0.058) | 0.198 (0.134, 0.267) | 0.099 (0.064, 0.136) | 0.094 (0.037, 0.150) | 0.065 (0.033, 0.099) |
| | P-value | <0.001 | <0.001 | 0.710 | <0.001 | <0.001 | 0.002 | <0.001 |
| Reader 10 | Without DeepFAN | 0.573 (0.521, 0.625) | 0.583 (0.538, 0.633) | 0.469 (0.405, 0.537) | 0.707 (0.649, 0.765) | 0.635 (0.561, 0.705) | 0.551 (0.495, 0.607) | 0.539 (0.486, 0.601) |
| | With DeepFAN | 0.707 (0.660, 0.754) | 0.706 (0.667, 0.745) | 0.631 (0.567, 0.688) | 0.788 (0.738, 0.843) | 0.764 (0.704, 0.827) | 0.663 (0.611, 0.713) | 0.691 (0.636, 0.735) |
| | Difference | 0.134 (0.086, 0.183) | 0.123 (0.078, 0.166) | 0.162 (0.094, 0.225) | 0.081 (0.029, 0.136) | 0.129 (0.077, 0.182) | 0.112 (0.073, 0.152) | 0.152 (0.097, 0.206) |
| | P-value | <0.001 | <0.001 | <0.001 | 0.006 | <0.001 | <0.001 | <0.001 |
| Reader 11 | Without DeepFAN | 0.681 (0.633, 0.729) | 0.670 (0.631, 0.713) | 0.680 (0.622, 0.738) | 0.658 (0.596, 0.714) | 0.683 (0.628, 0.738) | 0.655 (0.591, 0.712) | 0.682 (0.633, 0.732) |
| | With DeepFAN | 0.758 (0.715, 0.801) | 0.741 (0.706, 0.784) | 0.705 (0.651, 0.762) | 0.779 (0.724, 0.833) | 0.776 (0.722, 0.830) | 0.709 (0.650, 0.764) | 0.739 (0.691, 0.783) |
| | Difference | 0.077 (0.040, 0.113) | 0.071 (0.035, 0.112) | 0.025 (-0.022, 0.071) | 0.122 (0.064, 0.183) | 0.093 (0.049, 0.138) | 0.054 (0.015, 0.093) | 0.057 (0.020, 0.095) |
| | P-value | <0.001 | <0.001 | 0.377 | <0.001 | <0.001 | 0.002 | 0.004 |
| Reader 12 | Without DeepFAN | 0.773 (0.731, 0.815) | 0.730 (0.689, 0.771) | 0.954 (0.927, 0.979) | 0.486 (0.421, 0.555) | 0.669 (0.617, 0.720) | 0.908 (0.852, 0.957) | 0.786 (0.750, 0.819) |
| | With DeepFAN | 0.820 (0.782, 0.858) | 0.734 (0.695, 0.775) | 0.934 (0.903, 0.964) | 0.518 (0.452, 0.586) | 0.678 (0.628, 0.728) | 0.878 (0.819, 0.931) | 0.785 (0.749, 0.824) |
| | Difference | 0.047 (0.013, 0.082) | 0.004 (-0.028, 0.035) | -0.021 (-0.050, 0.008) | 0.032 (-0.024, 0.085) | 0.009 (-0.016, 0.035) | -0.030 (-0.082, 0.019) | -0.001 (-0.024, 0.020) |
| | P-value | 0.008 | 0.888 | 0.267 | 0.324 | 0.444 | 0.274 | 0.918 |

Numbers are values of diagnostic performance measures with 95% confidence intervals in parentheses. P values were calculated using the DeLong test for AUC, McNemar's test for sensitivity, specificity and accuracy, and nonparametric bootstrapping (1,000 iterations) for NPV, PPV and F1-score.

*Abbreviations:* AUC = area under curve, PPV = positive predictive value, NPV = negative predictive value.

**Supplementary Table 10. Stratified analysis of diagnostic performance among DeepFAN, unassisted readers, and AI-assisted readers**

| Stratification | AUC | | | Sensitivity | | | Specificity | | |
|---|---|---|---|---|---|---|---|---|---|
| | DeepFAN | Unassisted reader | AI-assisted reader | DeepFAN | Unassisted reader | AI-assisted reader | DeepFAN | Unassisted reader | AI-assisted reader |
| **Patient characteristics[^]** | | | | | | | | | |
| *Age (year)* | | | | | | | | | |
| ≥18 to ≤45 (n=32) | 0.976 (0.933, 1.000) | 0.660 (0.605, 0.716) | 0.801 (0.755, 0.846) | 0.929 (0.685, 0.996) | 0.654 (0.579, 0.729) | 0.788 (0.724, 0.853) | 0.842 (0.624, 0.945) | 0.601 (0.537, 0.664) | 0.728 (0.670, 0.786) |
| >45 to ≤60 (n=196) | 0.960 (0.935, 0.984) | 0.688 (0.667, 0.709) | 0.812 (0.795, 0.829) | 0.989 (0.941, 0.999) | 0.721 (0.695, 0.747) | 0.820 (0.797, 0.842) | 0.827 (0.743, 0.888) | 0.571 (0.544, 0.599) | 0.703 (0.677, 0.728) |
| >60 (n=172) | 0.964 (0.940, 0.988) | 0.755 (0.734, 0.776) | 0.842 (0.824, 0.859) | 0.969 (0.914, 0.990) | 0.810 (0.788, 0.833) | 0.851 (0.831, 0.872) | 0.851 (0.753, 0.915) | 0.555 (0.522, 0.588) | 0.703 (0.673, 0.733) |
| *Gender* | | | | | | | | | |
| Male (n=201) | 0.973 (0.954, 0.991) | 0.724 (0.704, 0.744) | 0.832 (0.816, 0.848) | 0.966 (0.906, 0.988) | 0.793 (0.769, 0.817) | 0.838 (0.816, 0.860) | 0.902 (0.833, 0.944) | 0.537 (0.511, 0.564) | 0.705 (0.681, 0.730) |
| Female (n=199) | 0.951 (0.923, 0.979) | 0.732 (0.712, 0.752) | 0.827 (0.810, 0.844) | 0.991 (0.952, 1.000) | 0.734 (0.711, 0.757) | 0.829 (0.809, 0.849) | 0.753 (0.652, 0.832) | 0.609 (0.579, 0.639) | 0.705 (0.677, 0.733) |
| **Nodule characteristics** | | | | | | | | | |
| *Nodule diameter (mm)* | | | | | | | | | |
| ≥4 to ≤10 (n=170) | 0.933 (0.890, 0.976) | 0.630 (0.605, 0.655) | 0.745 (0.723, 0.768) | 0.897 (0.815, 0.962) | 0.429 (0.397, 0.464) | 0.576 (0.542, 0.609) | 0.873 (0.806, 0.935) | 0.746 (0.721, 0.770) | 0.831 (0.809, 0.851) |
| >10 to ≤20 (n=219) | 0.946 (0.918, 0.974) | 0.656 (0.635, 0.677) | 0.769 (0.750, 0.787) | 0.960 (0.909, 0.983) | 0.766 (0.745, 0.788) | 0.811 (0.791, 0.831) | 0.800 (0.709, 0.868) | 0.478 (0.449, 0.507) | 0.632 (0.604, 0.660) |
| >20 to ≤30 (n=74) | 0.998 (0.994, 1.000) | 0.764 (0.731, 0.798) | 0.889 (0.865, 0.912) | 1.000 (0.927, 1.000) | 0.872 (0.845, 0.899) | 0.930 (0.910, 0.951) | 0.960 (0.804, 0.997) | 0.510 (0.453, 0.567) | 0.697 (0.645, 0.749) |
| *Nodule density* | | | | | | | | | |
| SN (n=192) | 0.977 (0.949, 1.000) | 0.725 (0.701, 0.748) | 0.816 (0.795, 0.837) | 0.881 (0.800, 0.962) | 0.725 (0.691, 0.757)[a] | 0.751 (0.719, 0.782)[a] | 0.962 (0.926, 0.992) | 0.626 (0.602, 0.649) | 0.783 (0.763, 0.802) |
| PSN (n=168) | 0.919 (0.878, … | 0.646 (0.619, 0.705) | 0.729 (0.705, … | 0.992 (0.975, … | 0.770 (0.748, … | 0.847 (0.829, … | 0.528 (0.357, … | 0.398 (0.352, … | 0.477 (0.431, … |
| GGN (n=103) | 0.905 (0.844, 0.966) | 0.622 (0.590, 0.653) | 0.729 (0.701, 0.757) | 0.920 (0.841, 0.982) | 0.452 (0.412, 0.492) | 0.585 (0.546, 0.628) | 0.792 (0.678, 0.900)[d] | 0.692 (0.654, 0.727) | 0.770 (0.735, 0.802)[d] |
| **Nodule location** | | | | | | | | | |
| RUL (n=125) | 0.962 (0.929, 0.996) | 0.662 (0.635, 0.689) | 0.772 (0.749, 0.795) | 0.970 (0.928, 1.000) | 0.682 (0.649, 0.714) | 0.767 (0.739, 0.796) | 0.810 (0.698, 0.904) | 0.575 (0.537, 0.611) | 0.695 (0.663, 0.728) |
| RML (n=36) | 0.935 (0.851, 1.000) | 0.678 (0.629, 0.728) | 0.777 (0.734, 0.821) | 0.842 (0.682, 1.000) | 0.623 (0.557, 0.682) | 0.728 (0.670, 0.783) | 0.882 (0.706, 1.000)[e] | 0.667 (0.595, 0.728) | 0.794 (0.731, 0.847)[e] |
| RLL (n=116) | 0.947 (0.905, 0.990) | 0.719 (0.692, 0.746) | 0.796 (0.772, 0.819) | 0.948 (0.887, 1.000) | 0.713 (0.679, 0.747) | 0.769 (0.735, 0.801) | 0.810 (0.704, 0.909) | 0.602 (0.565, 0.636) | 0.698 (0.663, 0.729) |
| LUL (n=99) | 0.928 (0.877, 0.979) | 0.725 (0.696, 0.753) | 0.819 (0.796, 0.842) | 0.948 (0.881, 1.000) | 0.741 (0.710, 0.775) | 0.792 (0.760, 0.823) | 0.854 (0.739, 0.955) | 0.612 (0.569, 0.658) | 0.752 (0.715, 0.790) |
| LLL (n=87) | 0.981 (0.950, 1.000) | 0.665 (0.632, 0.699) | 0.811 (0.784, 0.838) | 0.974 (0.909, 1.000) | 0.643 (0.599, 0.687) | 0.759 (0.718, 0.795) | 0.938 (0.857, 1.000) | 0.616 (0.576, 0.654) | 0.771 (0.739, 0.804) |
| **Diagnostic difficulty*** | | | | | | | | | |
| Low (n=267) | 0.994 (0.988, 1.000) | 0.913 (0.903, 0.923) | 0.956 (0.949, 0.963) | 0.987 (0.954, 0.996) | 0.868 (0.852, 0.883) | 0.908 (0.894, 0.921) | 0.964 (0.912, 0.986) | 0.838 (0.818, 0.858) | 0.910 (0.895, 0.925) |
| Intermediate (n=133) | 0.942 (0.907, 0.977) | 0.447 (0.419, 0.457) | 0.642 (0.615, 0.669) | 0.966 (0.883, 0.99) | 0.467 (0.430, 0.504) | 0.631 (0.595, 0.667) | 0.813 (0.711, 0.885) | 0.461 (0.429, 0.494) | 0.643 (0.612, 0.675) |
| High (n=63) | 0.644 (0.506, 0.781) | 0.113 (0.090, 0.136) | 0.229 (0.196, 0.262) | 0.714 (0.529, 0.847) | 0.190 (0.148, 0.232)[f] | 0.289 (0.240, 0.337)[f] | 0.571 (0.409, 0.720) | 0.167 (0.131, 0.202) | 0.343 (0.297, 0.388) |
| **Reader characteristics** | | | | | | | | | |
| Clinical trial center† | | | | | | | | | |
| Center I (n=3) | NA | 0.751 (0.727, 0.776) | 0.861 (0.842, 0.880) | NA | 0.811 (0.782, 0.839) | 0.885 (0.862, 0.908) | NA | 0.553 (0.515, 0.590) | 0.721 (0.687, 0.755) |
| Center II (n=4) | NA | 0.669 (0.644, 0.693) | 0.781 (0.760, 0.802) | NA | 0.646 (0.616, 0.676) | 0.743 (0.715, 0.770) | NA | 0.599 (0.567, 0.631) | 0.734 (0.705, 0.763) |
| Center III (n=5) | NA | 0.692 (0.671, 0.714) | 0.779 (0.760, 0.798) | NA | 0.659 (0.632, 0.686) | 0.720 (0.695, 0.746) | NA | 0.641 (0.612, 0.669) | 0.733 (0.707, 0.759) |
| Experience (year) | | | | | | | | | |
| 1-2 (n=6) | NA | 0.720 (0.702, 0.741) | 0.777 (0.761, 0.799) | NA | 0.698 (0.674, 0.723) | 0.795 (0.774, 0.815) | NA | 0.601 (0.580, 0.632) | 0.731 (0.707, 0.754) |
| 3-5 (n=6) | NA | 0.690 (0.676, 0.716) | 0.831 (0.795, 0.848) | NA | 0.687 (0.664, 0.710) | 0.743 (0.721, 0.766) | NA | 0.604 (0.578, 0.628) | 0.730 (0.664, 0.710) |
| Education | | | | | | | | | |
| BM (n=4) | NA | 0.679 (0.655, 0.703) | 0.784 (0.763, 0.804) | NA | 0.667 (0.637, 0.697) | 0.762 (0.736, 0.789) | NA | 0.619 (0.587, 0.651) | 0.682 (0.652, 0.713) |
| MM (n=5) | NA | 0.683 (0.662, 0.704) | 0.780 (0.761, 0.799) | NA | 0.642 (0.615, 0.669) | 0.705 (0.679, 0.730) | NA | 0.624 (0.596, 0.653) | 0.775 (0.750, 0.799) |
| MD (n=3) | NA | 0.751 (0.727, 0.776) | 0.861 (0.842, 0.880) | NA | 0.811 (0.782, 0.839) | 0.885 (0.862, 0.908) | NA | 0.553 (0.515, 0.590) | 0.721 (0.687, 0.755) |

Numbers are values of diagnostic performance measures with 95% confidence intervals in parentheses. Unless otherwise, pairwise comparisons among AI predictions, unassisted predictions, and AI-assisted predictions over each subgroup showed a *P*-value of less than 0.01 for all performance measures. The DeLong test was used to compare AUCs between paired groups, while McNemar's test was employed to assess differences in sensitivity and specificity. For comparisons involving more than two groups, Bonferroni correction was applied to adjust for multiple testing.

^ The patient-level analysis results were reported in groups stratified by patient characteristics. All other results were reported using nodule-level results.

\* The diagnostic difficulty of a pulmonary nodule was defined as low, intermediate, and high when more than two-thirds, between one-third and two-thirds, and less than one-third, respectively, of unassisted readers correctly classified the nodule as benign or malignant.

† Clinical trial center I, II, and III represent Peking University People's Hospital, Wuhan Third Hospital, and Huangshi Central Hospital, respectively.

a *P* value for sensitivity comparison over the SN subgroup between unassisted readers and AI-assisted readers is 0.333.

b *P* value for specificity comparison over the PSN subgroup between DeepFAN and AI-assisted readers is 0.212.

c *P* value for specificity comparison over the PSN subgroup between unassisted readers and AI-assisted readers is 0.016.

d *P* value for specificity comparison over the GGN subgroup between DeepFAN and AI-assisted readers is 0.619.

e *P* value for specificity comparison over the RML subgroup between DeepFAN and AI-assisted readers is 0.023.

f *P* value for sensitivity comparison over the high diagnostic difficulty subgroup between unassisted readers and AI-assisted readers is 0.016.

*Abbreviations:* AUC = area under curve, SN = solid nodule, PSN = part-solid nodule, GGN = ground-glass nodule, RUL = right upper lobe, RML = right middle lobe, RLL = right lower lobe, LUL = left upper lobe, LLL = left lower lobe, NA = not applicable, BM = bachelor of medicine, MM = master of medicine, MD = doctor of medicine.

**Supplementary Table 11. Diagnostic performance improvement across stratified subgroups: unassisted vs. AI-assisted readers.**

| Stratification | ΔAUC | P value | ΔSensitivity | P value | ΔSpecificity | P value |
|---|---|---|---|---|---|---|
| **Patient characteristics^** | | | | | | |
| Age (year) | | | | | | |
| ≥18 to ≤45 (n=32) | 0.140 (0.099, 0.186) | <0.001 | 0.135 (0.073, 0.201) | 0.002 | 0.127 (0.074, 0.186) | 0.001 |
| >45 to ≤60 (n=196) | 0.124 (0.108, 0.141) | <0.001 | 0.099 (0.073, 0.124) | <0.001 | 0.131 (0.106, 0.156) | <0.001 |
| >60 (n=172) | 0.087 (0.070, 0.105) | <0.001 | 0.041 (0.018, 0.062) | 0.002 | 0.148 (0.117, 0.179) | <0.001 |
| Gender | | | | | | |
| Male (n=201) | 0.108 (0.093, 0.124) | <0.001 | 0.045 (0.023, 0.068) | 0.001 | 0.168 (0.142, 0.193) | <0.001 |
| Female (n=199) | 0.095 (0.079, 0.112) | <0.001 | 0.095 (0.073, 0.118) | <0.001 | 0.096 (0.069, 0.123) | <0.001 |
| **Nodule characteristics** | | | | | | |
| Nodule diameter (mm) | | | | | | |
| ≥4 to ≤10 (n=170) | 0.115 (0.097, 0.134) | <0.001 | 0.147 (0.114, 0.177) | <0.001 | 0.085 (0.062, 0.106) | <0.001 |
| >10 to ≤20 (n=219) | 0.113 (0.096, 0.129) | <0.001 | 0.045 (0.024, 0.064) | <0.001 | 0.154 (0.125, 0.182) | <0.001 |
| >20 to ≤30 (n=74) | 0.125 (0.098, 0.152) | <0.001 | 0.058 (0.033, 0.085) | <0.001 | 0.187 (0.132, 0.243) | <0.001 |
| Nodule density | | | | | | |
| SN (n=192) | 0.091 (0.075, 0.107) | <0.001 | 0.027 (-0.003, 0.058) | 0.333 | 0.157 (0.136, 0.181) | <0.001 |
| PSN (n=168) | 0.083 (0.061, 0.104) | <0.001 | 0.077 (0.056, 0.097) | <0.001 | 0.079 (0.033, 0.126) | 0.016 |
| GGN (n=103) | 0.108 (0.086, 0.129) | <0.001 | 0.133 (0.099, 0.166) | <0.001 | 0.079 (0.048, 0.107) | <0.001 |
| Nodule location | | | | | | |
| RUL (n=125) | 0.110 (0.089, 0.131) | <0.001 | 0.086 (0.058, 0.116) | <0.001 | 0.121 (0.088, 0.157) | <0.001 |
| RML (n=36) | 0.099 (0.063, 0.133) | <0.001 | 0.105 (0.048, 0.167) | 0.002 | 0.127 (0.072, 0.181) | <0.001 |
| RLL (n=116) | 0.077 (0.057, 0.095) | <0.001 | 0.056 (0.027, 0.085) | 0.007 | 0.096 (0.067, 0.129) | <0.001 |
| LUL (n=99) | 0.094 (0.073, 0.113) | <0.001 | 0.050 (0.022, 0.077) | 0.006 | 0.140 (0.102, 0.177) | <0.001 |
| LLL (n=87) | 0.146 (0.120, 0.171) | <0.001 | 0.115 (0.074, 0.156) | <0.001 | 0.155 (0.116, 0.191) | <0.001 |
| Diagnostic difficulty* | | | | | | |
| Low (n=267) | 0.043 (0.034, 0.052) | <0.001 | 0.040 (0.024, 0.056) | <0.001 | 0.072 (0.053, 0.092) | <0.001 |
| Intermediate (n=133) | 0.195 (0.167, 0.220) | <0.001 | 0.164 (0.122, 0.201) | <0.001 | 0.182 (0.149, 0.217) | <0.001 |
| High (n=63) | 0.116 (0.088, 0.144) | <0.001 | 0.098 (0.052, 0.146) | 0.016 | 0.176 (0.133, 0.223) | <0.001 |

## Reader characteristics

| | | | | | |
|---|---|---|---|---|---|
| Clinical trial center[†] | | | | | |
| Center I (n=3) | 0.110 (0.089, 0.129) | <0.001 | 0.075 (0.047, 0.101) | <0.001 | 0.168 (0.133, 0.204) | <0.001 |
| Center II (n=4) | 0.112 (0.093, 0.130) | <0.001 | 0.096 (0.069, 0.124) | <0.001 | 0.135 (0.107, 0.163) | <0.001 |
| Center III (n=5) | 0.087 (0.072, 0.101) | <0.001 | 0.061 (0.041, 0.084) | <0.001 | 0.093 (0.066, 0.115) | <0.001 |
| Experience (year) | | | | | |
| 1-2 (n=6) | 0.102 (0.087, 0.115) | <0.001 | 0.097 (0.076, 0.117) | <0.001 | 0.125 (0.100, 0.148) | <0.001 |
| 3-5 (n=6) | 0.105 (0.089, 0.120) | <0.001 | 0.056 (0.033, 0.079) | <0.001 | 0.126 (0.103, 0.150) | <0.001 |
| Education | | | | | |
| BM (n=4) | 0.105 (0.089, 0.122) | <0.001 | 0.095 (0.073, 0.121) | <0.001 | 0.063 (0.036, 0.090) | <0.001 |
| MM (n=5) | 0.097 (0.082, 0.114) | <0.001 | 0.062 (0.037, 0.088) | <0.001 | 0.150 (0.124, 0.177) | <0.001 |
| MD (n=3) | 0.110 (0.090, 0.130) | <0.001 | 0.075 (0.048, 0.102) | <0.001 | 0.168 (0.131, 0.210) | <0.001 |

Numbers are values of improvement in diagnostic performance measures with 95% confidence intervals in parentheses. Confidence intervals were estimated using nonparametric bootstrap with 1,000 iterations. P values were calculated using either DeLong test or McNemar's test as specified in Supplementary Table 10.

[†] Clinical trial center I, II, and III represent Peking University People's Hospital, Wuhan Third Hospital, and Huangshi Central Hospital, respectively.

^ Patient-level analysis results were presented stratified by patient characteristics, while all other findings were reported at the nodule level.

* The diagnostic difficulty of a pulmonary nodule was defined as low, intermediate, and high when more than two-thirds, between one-third and two-thirds, and less than one-third, respectively, of unassisted readers correctly classified the nodule as benign or malignant.

*Abbreviations:* AUC = area under curve, SN = solid nodule, PSN = part-solid nodule, GGN = ground-glass nodule, RUL = right upper lobe, RML = right middle lobe, RLL = right lower lobe, LUL = left upper lobe, LLL = left lower lobe, NA = not applicable, BM = bachelor of medicine, MM = master of medicine, MD = doctor of medicine.

**Supplementary Table 12. Logistic regression analysis of nodule features and AI predicted malignancy in the clinical trial**

| Variable | Univariable | | Multivariable | |
|---|---|---|---|---|
| | OR (95% CI) | *P* value | OR (95% CI) | *P* value |
| Nodule diameter (unit: mm) | 1.09 (1.05,1.13) | **<0.001** | 1.11 (1.06,1.17) | **<0.001** |
| Nodule density (reference: solid nodule) | | | | |
| Part-solid nodule | 17.53 (10.01,30.69) | **<0.001** | 30.05 (14.96,60.38) | **<0.001** |
| Ground-glass nodule | 2.94 (1.78,4.82) | **<0.001** | 18.05 (8.23,39.57) | **<0.001** |
| Location (reference: right upper lobe) | | | | |
| Right middle lobe | 0.65 (0.31,1.36) | 0.249 | 0.97 (0.36,2.62) | 0.959 |
| Right lower lobe | 0.85 (0.51,1.42) | 0.538 | 1.34 (0.68,2.63) | 0.401 |
| Left upper lobe | 1.04 (0.60,1.78) | 0.901 | 1.21 (0.60,2.46) | 0.597 |
| Left lower lobe | 0.58 (0.33,1.00) | 0.050 | 0.70 (0.33,1.46) | 0.339 |
| Spiculation (reference: no) | 3.69 (2.50,5.43) | **<0.001** | 4.67 (2.51,8.71) | **<0.001** |
| Lobulation (reference: no) | 6.84 (3.68,12.72) | **<0.001** | 4.26 (1.95,9.28) | **<0.001** |

*Abbreviations:* OR = odds ratio, CI = confidence interval.

**Supplementary Table 13. Univariable and multivariable generalized linear mixed analyses for factors influencing the accuracy of AI-assisted reading**

| Variable | Univariable | | Multivariable | |
|---|---|---|---|---|
| | β value | *P* value | β value | *P* value |
| **Diagnostic related results** | | | | |
| Pathology (reference: benign) | 0.56 | **0.002** | NA | NA |
| Correct AI suggestions (reference: incorrect) | 3.01 | **<0.001** | 1.72 | **<0.001** |
| Correct reading at first session (reference: incorrect) [*] | 2.88 | **<0.001** | NA | NA |
| Interaction between correct AI suggestions and correct reading[#] | -0.71 | **0.036** | NA | NA |
| **Patient characteristics** | | | | |
| Clinical trial center (reference: center I) | | | | |
|    Center II | -0.39 | 0.189 | | |
|    Center III | -0.35 | 0.065 | | |
| Patient age (unit: year) | 0.01 | 0.134 | | |
| Patient sex (reference: female) | 0.04 | 0.802 | | |
| **Nodule characteristics** | | | | |
| Nodule diameter (unit: mm) | 0.04 | **0.010** | 0.00 | 0.631 |
| Nodule density (reference: solid nodule) | | | | |
|    Part-solid nodule | 0.07 | 0.718 | 0.10 | 0.378 |
|    Ground-glass nodule | -0.62 | **0.008** | -0.11 | 0.416 |
| Nodule location (reference: left lung) | -0.28 | 0.128 | | |
| Spiculation (reference: no) | 0.18 | 0.334 | | |
| Lobulation (reference: no) | -0.70 | **0.010** | -0.20 | 0.197 |
| Diagnostic difficulty (reference: low) [$] | | | | |
|    Intermediate | -1.86 | **<0.001** | -1.65 | **<0.001** |
|    High | -3.32 | **<0.001** | -2.68 | **<0.001** |
| **CT image characteristics** | | | | |
| Slice thickness ≥1 mm (reference: <1 mm) | 0.30 | 0.091 | | |
| **Reader characteristics** | | | | |
| Clinical trial center (reference: center I) | | | | |
|    Center II | -0.54 | **0.031** | NA | NA |
|    Center III | -0.62 | **0.010** | NA | NA |
| Reading experience of 3-5 (references: 1-2 [unit: year]) | -0.07 | 0.778 | | |
| Education (reference: doctor of medicine) | | | | |
|    Master of medicine | -0.52 | **0.028** | -0.53 | **0.034** |
|    Bachelor of medicine | -0.66 | **0.007** | -0.67 | **0.009** |
| Annual chest CT diagnoses ≥10,000 (reference:<10,000[unit: case]) | -0.22 | 0.374 | | |
| Research experience in medical imaging AI (reference: no) | -0.02 | 0.924 | | |
| Familiar with background knowledge of AI (reference: unfamiliar)^ | -0.02 | 0.924 | | |
| Attitude of trust toward AI (reference: neutral) [+] | -0.09 | 0.719 | | |
| Total grit score[%] | 0.03 | 0.080 | | |

*Correct reading at first session denotes the situation where the assessment from the first reading session (without AI) is the same as the ground truth. [#]A variable combination approach where the categorical variable of AI suggestion is multiplied by that of the correct reading at first session. [$]The diagnostic difficulty of a pulmonary

nodule was defined as low, intermediate, and high when more than two-thirds, between one-third and two-thirds, and less than one-third, respectively, of the unassisted readers correctly classified the nodule as benign or malignant. ^  The data were categorized into binary groups: "unfamiliar" (a rating score ≤ 3 for Question 8 in the questionnaire) and "familiar" (a rating score > 3). [+] The data automatically fell into binary groups: "neutral" (a rating score of 3 for Question 9 in the questionnaire) and "mostly believe" (a rating score of 4). [%]Total grit score for personality traits related to stability of interest and persistence of effort (Supplementary Table 14 and reader questionnaire).

*Abbreviations:* AI = artificial intelligence, NA = not applicable.

**Supplementary Table 14. Reader's personal characteristics, experience with medical imaging AI, attitude of trust toward AI and Grit score**

| Reader number | Workplace | Working experience | Education | Whether number of annual chest CT diagnoses exceeding 10,000 cases or not? | Do you have prior experience using AI tools for chest CT diagnosis? | Do you have research experience in medical imaging AI? | Rate your level of knowledge regarding medical imaging AI technology* | Rate your attitude of trust toward AI-based computer-assisted diagnosis systems# | Total Grit score& (points) |
|---|---|---|---|---|---|---|---|---|---|
| 01 | Center I | 2 years | MD | No | Yes | Yes | 4 | 4 | 50 |
| 02 | Center I | 1 year | MD | No | Yes | No | 3 | 3 | 46 |
| 03 | Center I | 1 year | MD | No | Yes | Yes | 4 | 4 | 37 |
| 04 | Center II | 5 years | MM | No | Yes | No | 3 | 4 | 39 |
| 05 | Center II | 2 years | BM | No | Yes | No | 3 | 3 | 44 |
| 06 | Center II | 3 years | MM | No | Yes | No | 3 | 3 | 46 |
| 07 | Center II | 4 years | MM | No | Yes | Yes | 4 | 4 | 51 |
| 08 | Center III | 1 year | MM | Yes | Yes | No | 3 | 4 | 48 |
| 09 | Center III | 2 years | MM | Yes | Yes | No | 3 | 4 | 34 |
| 10 | Center III | 4 years | BM | Yes | Yes | Yes | 4 | 4 | 36 |
| 11 | Center III | 4 years | BM | Yes | Yes | Yes | 5 | 3 | 36 |
| 12 | Center III | 5 years | BM | Yes | Yes | Yes | 4 | 4 | 43 |

Clinical trial centers I, II, and III represent Peking University People's Hospital, Wuhan Third Hospital, and Huangshi Central Hospital, respectively. * Level of knowledge regarding medical imaging AI technology was rated on a five-point scale from 1 ("Never heard of") to 5 ("Master skill of AI"), with higher scores reflecting greater expertise. # Attitude of trust toward AI-based computer-assisted diagnosis systems was rated on a five-point scale from 1 ("Not believe at all") to 5 ("Completely believe"), where higher scores indicate greater perceived belief. &Total Grit score reflects personality traits related to stability of interest and persistence of effort. Please refer to reader questionnaire for details.

*Abbreviations:* BM = bachelor of medicine, MM = master of medicine, MD = doctor of medicine.

**Reader Questionnaire**

Electronic questionnaires were administered to junior radiologists right before the multi-reader multi-case clinical trial. Question 1 to 13 characterize reader's personal characteristics, experience with medical imaging AI, and attitude of trust toward AI. Question 14 and 15 are components of Grit score, each consisting of six questions that individually describe interest stability and effort persistence. The total Grit score was calculated by summing the scores for the 12 questions. The questionnaires were originally presented in Chinese. For the reader's convenience, we have translated the Chinese text into English. The count and proportion for each question are also provided.

*Personal characteristics*

1.  Please fill in the blank with your name ______
2.  Please fill in the blank with your age ______ (years old)
3.  Please select your gender [single choice]

| Options | Counts (percentage%) |
| --- | --- |
| 1. Male | 4(33.33) |
| 2. Female | 8(66.67) |

4.  Please fill in the blank with your major for the Bachelor's degree ______
5.  How many months have you participated in clinical practice experience (not limited to the radiology department) prior to the start of China's standardized residency training? ______ (months)
6.  Please select your workplace [single choice]

| Options | Counts (percentage%) |
| --- | --- |
| 1. Peking University People's Hospital | 3(25) |
| 2.Wuhan Third Hospital | 4(33.33) |
| 3. Huangshi Central Hospital | 5(41.67) |

7.  How many years have you been working in diagnostic imaging at your work place? [single choice]

| Options | Counts (percentage%) |
|---|---|
| 1.  1 year | 3(25) |
| 2.  2 years | 3(25) |
| 3.  3 years | 1(8.33) |
| 4.  4 years | 3(25) |
| 5.  5 years | 2(16.67) |

8.  Please select your highest education level [single choice]

| Options | Counts (percentage%) |
|---|---|
| 1.  Doctor of Medicine | 3(25) |
| 2.  Master of Medicine | 5(41.67) |
| 3.  Bachelor of Medicine | 4(33.33) |

9.  What is your estimated number of annual chest CT diagnoses [Single Choice]

| Options | Counts (percentage%) |
|---|---|
| 1.  <10000 cases per year | 7(58.33) |
| 2.  ≥10000 cases per year | 5(41.67) |

*Experience with medical imaging AI*

10. Do you have prior experience using AI tools for chest CT diagnosis? [single choice]

| Options | Counts (percentage%) |
|---|---|
| 1. Yes | 12(100) |
| 2. No | 0(0) |

## 11. Do you have research experience in medical imaging AI? [single choice]

| Options | Counts (percentage%) |
|---|---|
| 1. Yes | 6(50) |
| 2. No | 6(50) |

## 12. Rate your level of knowledge regarding medical imaging AI technology

| Options | Counts (percentage%) |
|---|---|
| 1. Never heard of | 0(0) |
| 2. Have heard but knew little | 0(0) |
| 3. Have read related information from journals or books | 6(50) |
| 4. Familiar with AI | 5(41.67) |
| 5. Master skill of AI | 1(8.33) |

## 13. Rate your attitude of trust toward AI-based computer-assisted diagnosis systems

| Options | Counts (percentage%) |
|---|---|
| 1. Not believe at all | 0(0) |
| 2. Not much believe | 0(0) |

| | | |
|---|---|---|
| 3. Neutral | 4(33.33) | |
| 4. Mostly believe | 8(66.67) | |
| 5. Completely believe | 0(0) | |

*Grit score*

14. To better understand your interest stability, please choose the best option that best fit you for the following questions

| Questions | Not like me at all (1 point) | Not much like me (2 points) | Somewhat like me (3 points) | Mostly like me (4 points) | Very much like me (5 points) |
|---|---|---|---|---|---|
| 1. I often set a goal but later choose to pursue a different one. | 0(0%) | 8(66.67%) | 4(33.33%) | 0(0%) | 0(0%) |
| 2. New ideas and projects sometimes distract me from previous ones. | 0(0%) | 5(41.67%) | 5(41.67%) | 2(16.67%) | 0(0%) |
| 3. Every few months, I become interested in something new. | 1(8.33%) | 6(50%) | 5(41.67%) | 0(0%) | 0(0%) |
| 4. My interests change every year. | 3(25%) | 8(66.67%) | 1(8.33%) | 0(0%) | 0(0%) |
| 5. I was once fascinated by a certain idea or project for a while, but eventually lost interest. | 1(8.33%) | 5(41.67%) | 4(33.33%) | 2(16.67%) | 0(0%) |
| 6. I find it difficult to stay focused on projects that take more than a few months to complete. | 1(8.33%) | 9(75%) | 2(16.67%) | 0(0%) | 0(0%) |

15. To better understand your effort persistence, please choose the best option that best fit you for following question

| Questions | Not like me at all (1 point) | Not much like me (2 points) | Somewhat like me (3 points) | Mostly like me (4 points) | Very much like me (5 points) |
|---|---|---|---|---|---|
| 1. I have accomplished a goal that required years of effort to achieve. | 0(0%) | 4(33.33%) | 3(25%) | 2(16.67%) | 3(25%) |
| 2. I have overcome setbacks to complete an important challenge. | 1(8.33%) | 3(25%) | 4(33.33%) | 2(16.67%) | 2(16.67%) |
| 3. Whatever I start, I will see it through to the end. | 0(0%) | 2(16.67%) | 5(41.67%) | 5(41.67%) | 0(0%) |
| 4. Setbacks do not discourage me. | 0(0%) | 0(0%) | 8(66.67%) | 4(33.33%) | 0(0%) |
| 5. I am a hard-working person | 0(0%) | 1(8.33%) | 4(33.33%) | 3(25%) | 4(33.33%) |
| 6. I am a diligent person | 0(0%) | 2(16.67%) | 3(25%) | 5(41.67%) | 2(16.67%) |

**Extended Data Figure 1**

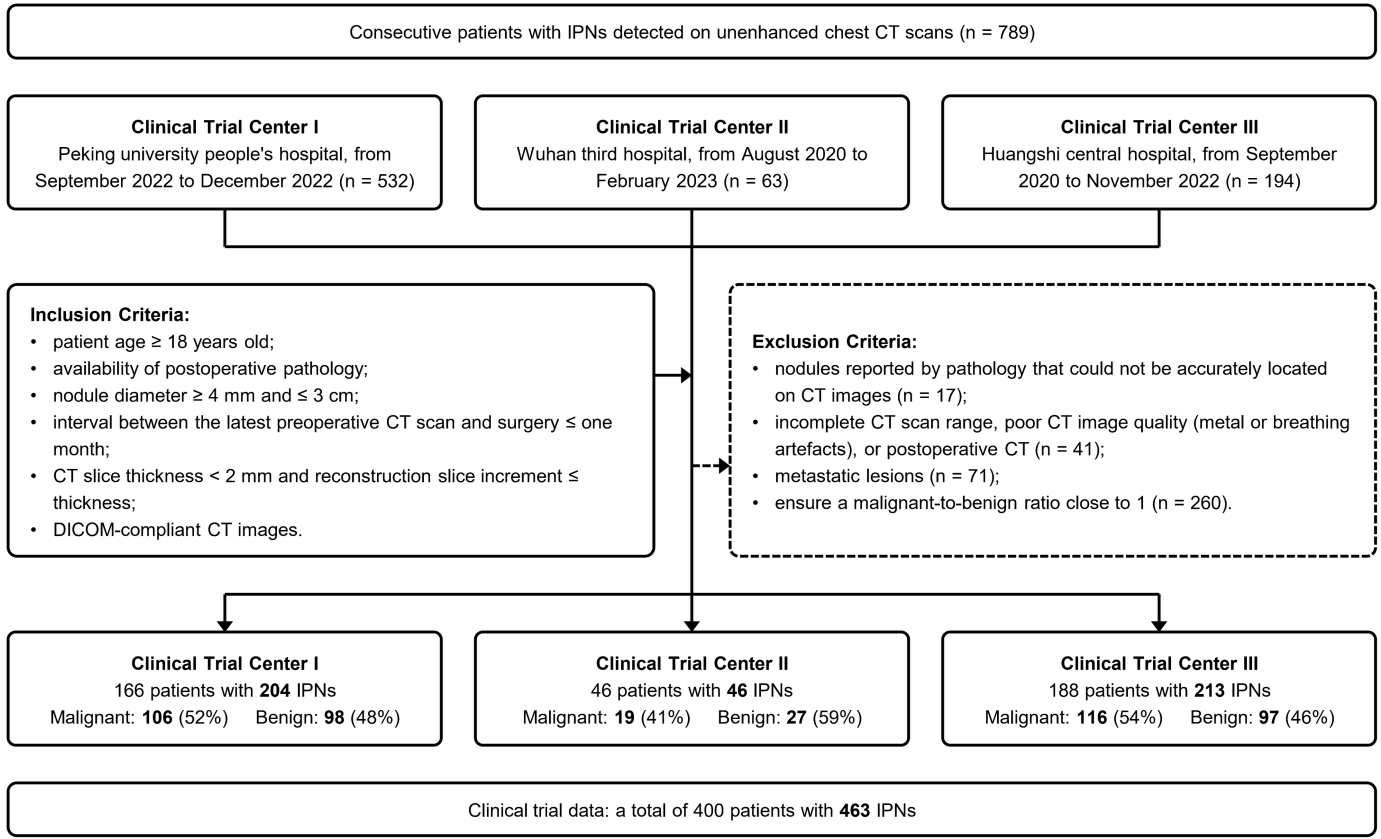

Consecutive patients with IPNs detected on unenhanced chest CT scans (n = 789)

**Clinical Trial Center I**
Peking university people's hospital, from September 2022 to December 2022 (n = 532)

**Clinical Trial Center II**
Wuhan third hospital, from August 2020 to February 2023 (n = 63)

**Clinical Trial Center III**
Huangshi central hospital, from September 2020 to November 2022 (n = 194)

**Inclusion Criteria:**
- patient age ≥ 18 years old;
- availability of postoperative pathology;
- nodule diameter ≥ 4 mm and ≤ 3 cm;
- interval between the latest preoperative CT scan and surgery ≤ one month;
- CT slice thickness < 2 mm and reconstruction slice increment ≤ thickness;
- DICOM-compliant CT images.

**Exclusion Criteria:**
- nodules reported by pathology that could not be accurately located on CT images (n = 17);
- incomplete CT scan range, poor CT image quality (metal or breathing artefacts), or postoperative CT (n = 41);
- metastatic lesions (n = 71);
- ensure a malignant-to-benign ratio close to 1 (n = 260).

**Clinical Trial Center I**
166 patients with **204** IPNs
Malignant: **106** (52%)    Benign: **98** (48%)

**Clinical Trial Center II**
46 patients with **46** IPNs
Malignant: **19** (41%)    Benign: **27** (59%)

**Clinical Trial Center III**
188 patients with **213** IPNs
Malignant: **116** (54%)    Benign: **97** (46%)

Clinical trial data: a total of 400 patients with **463** IPNs

**Extended Data Figure 2**

## Extended Data Figure 3

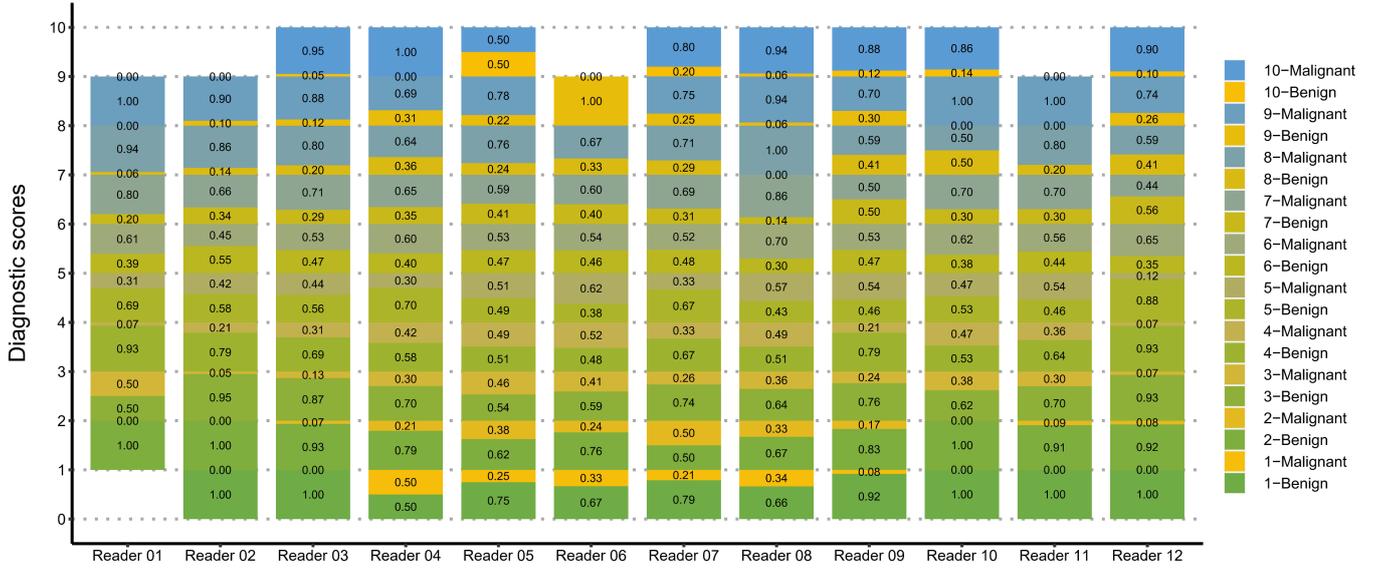

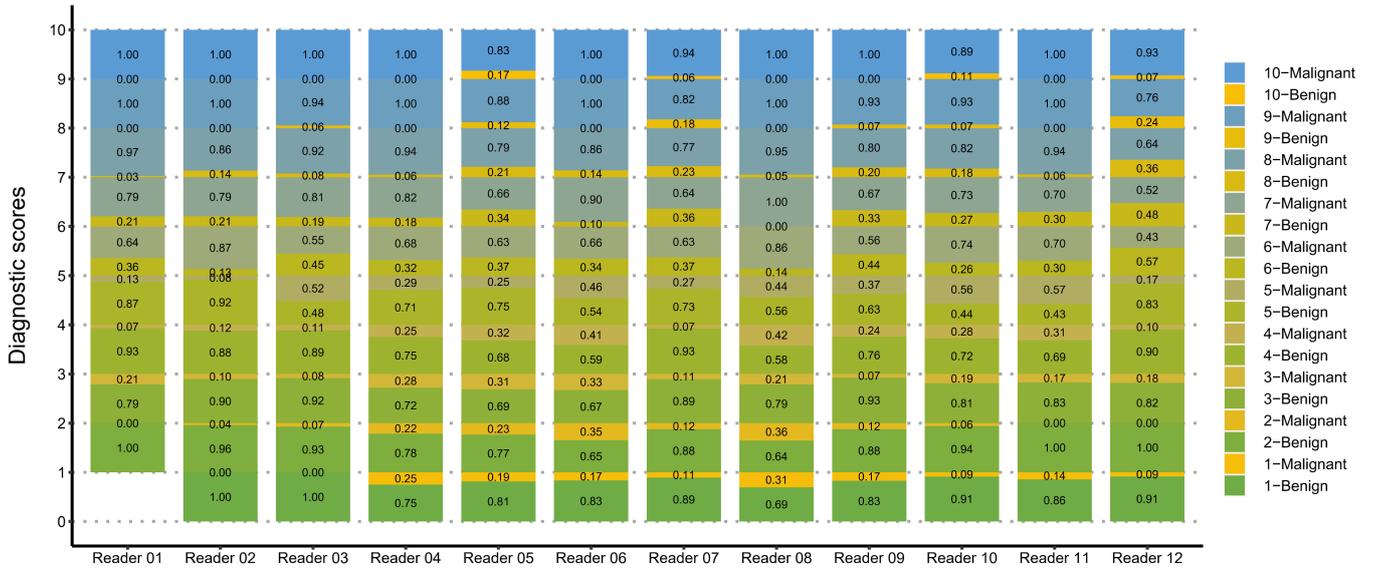

**Extended Data Figure 4**

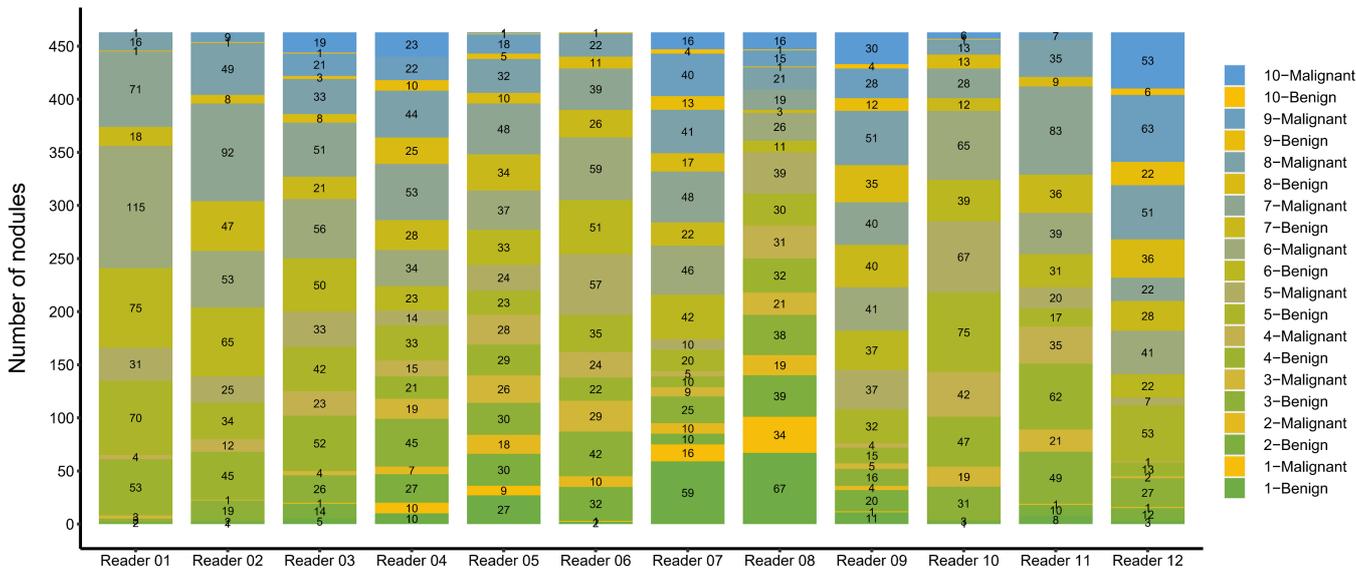

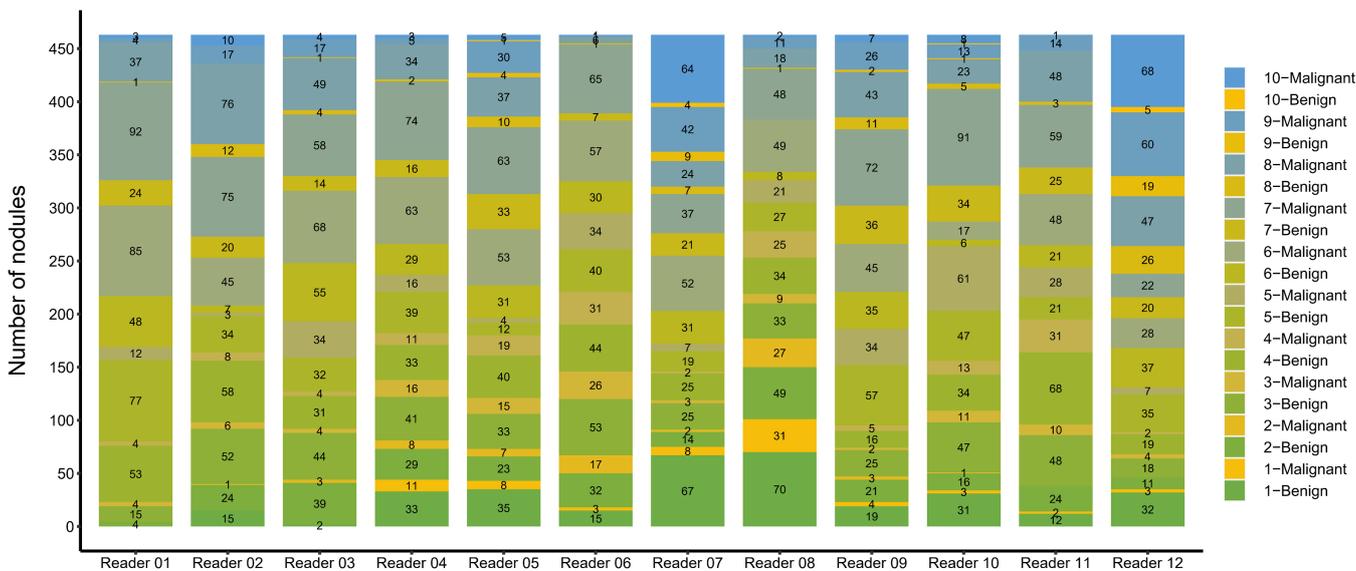

**Extended Data Figure 5**

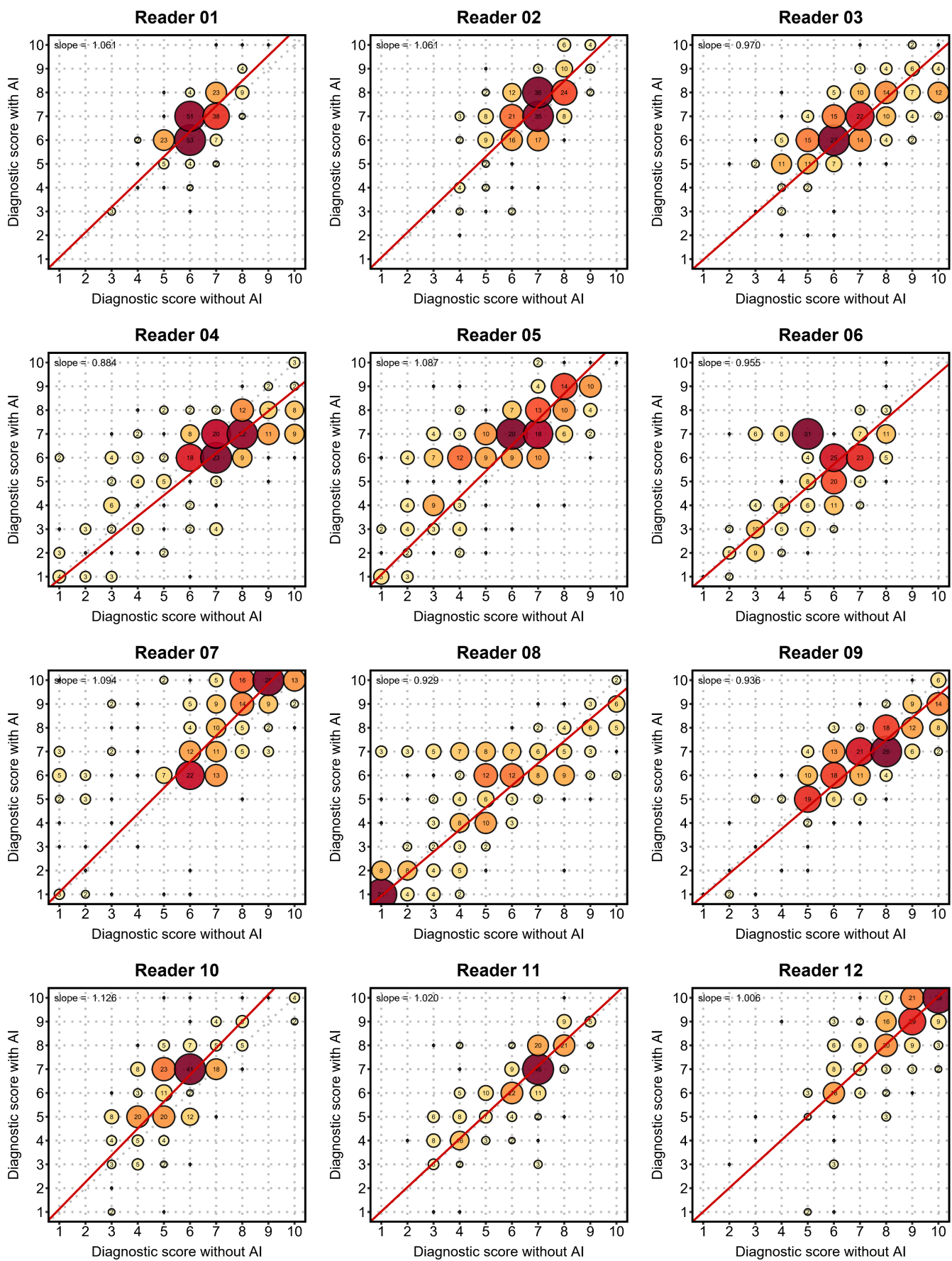

**Extended Data Figure 6**

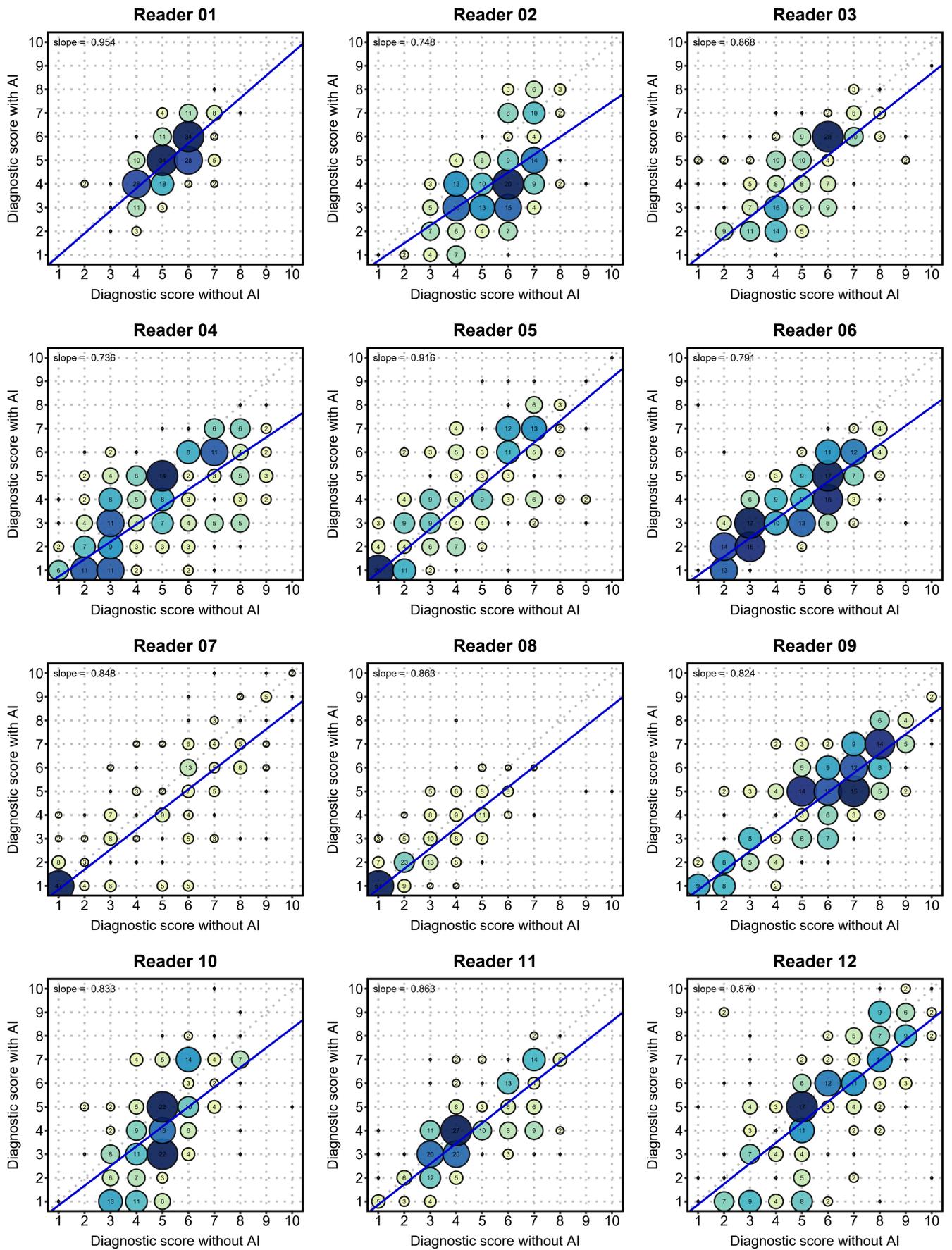

**Extended Data Figure 7**

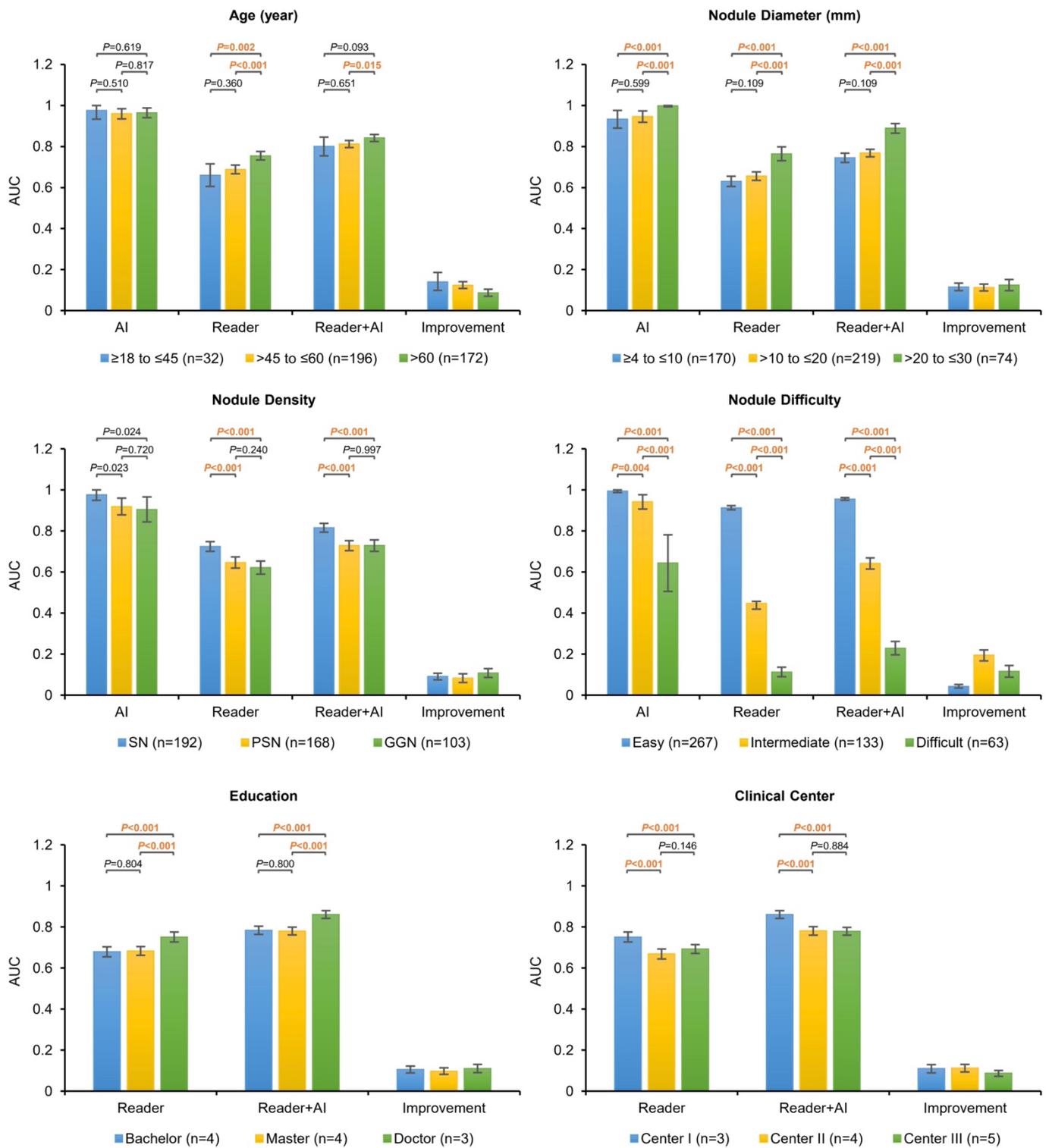

**Extended Data Figure 8**

### Twelve readers (n = 5556)

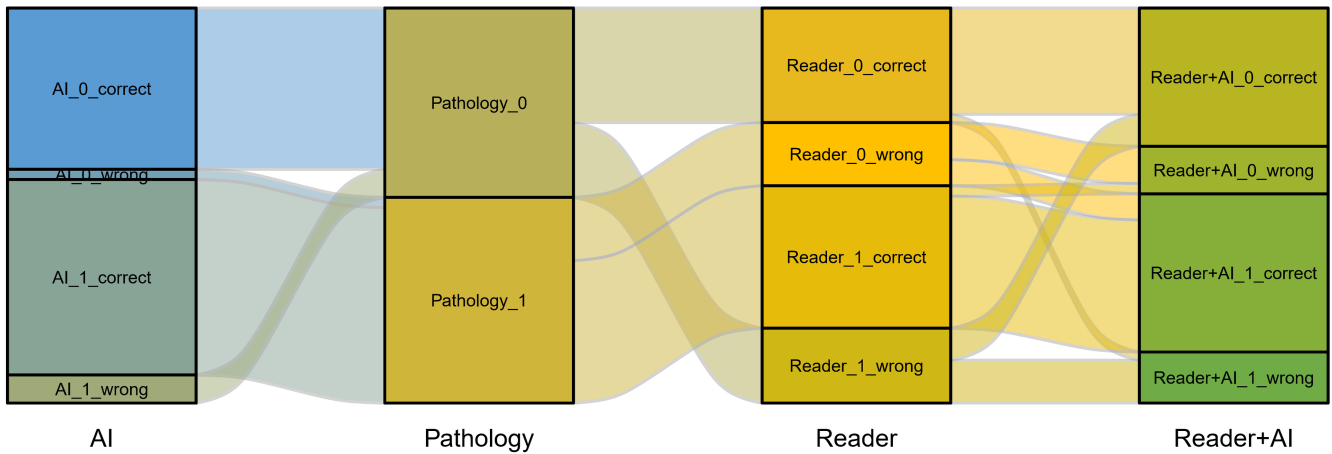

### Reader 02 (n = 463)

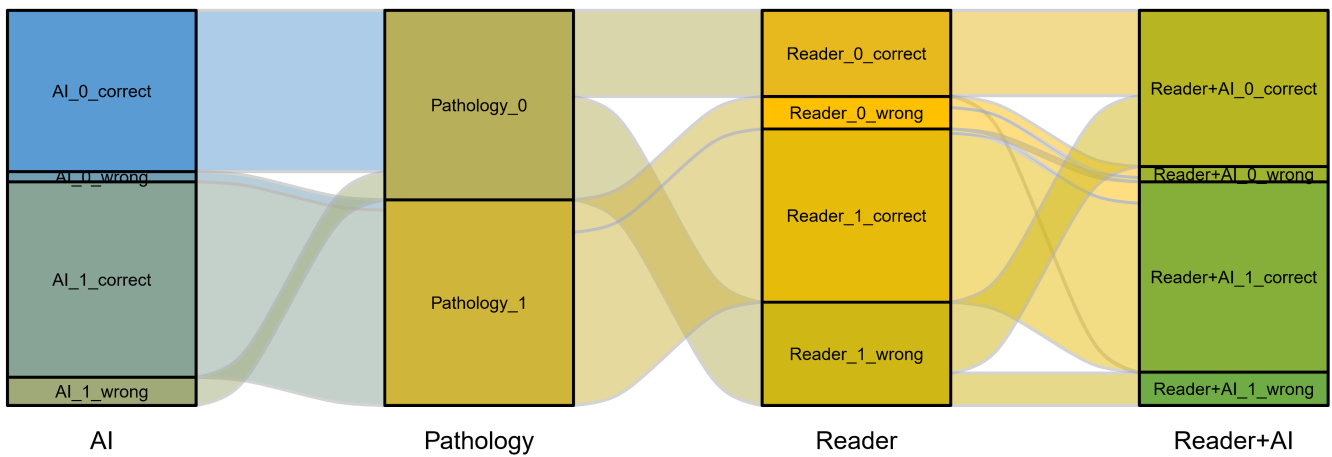

### Reader 12 (n = 463)

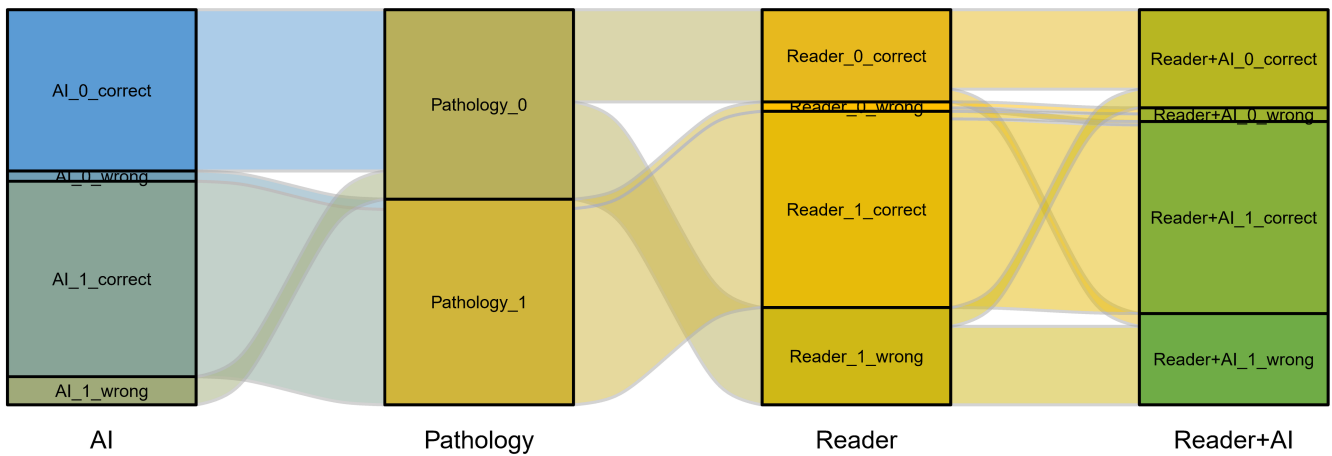

## Extended Data Figure 9

**(a) Nodules correctly classified by AI but misclassified by readers, and accurately revised in human-machine collaboration**

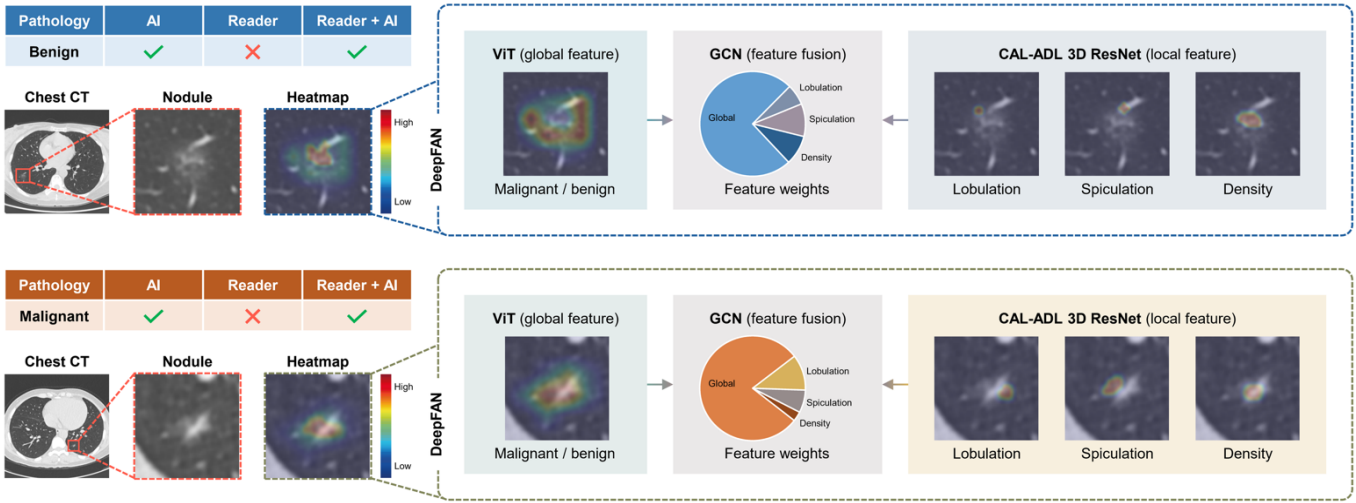

**(b) Nodules correctly classified by AI but misclassified by readers, and still misclassified in human-machine collaboration**

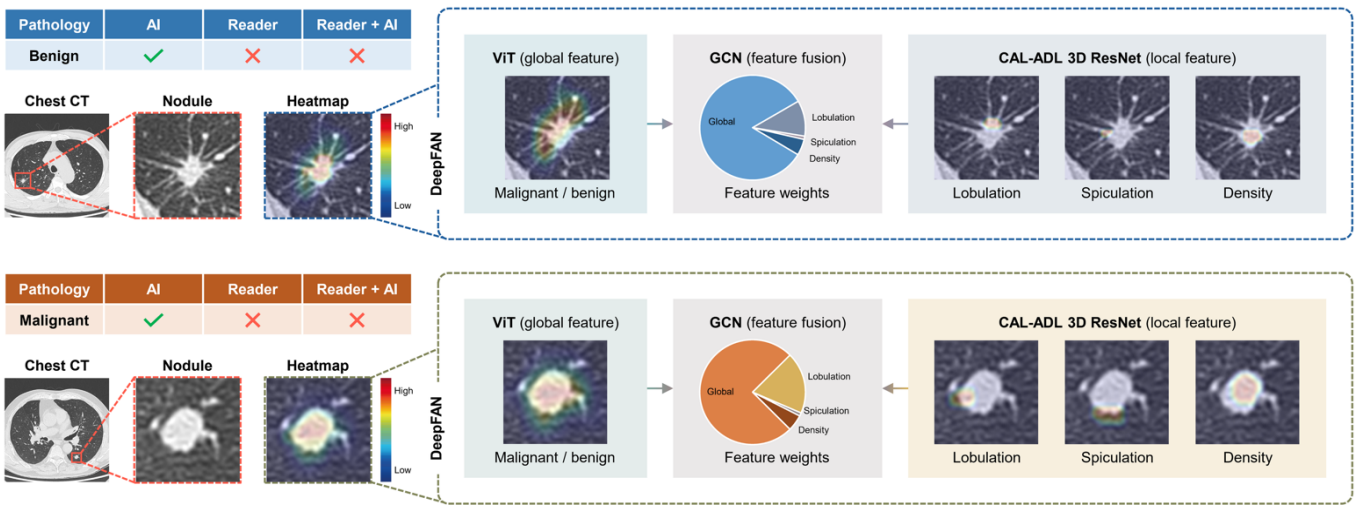

**(c) Nodules misclassified by AI, unassisted readers and human-machine collaboration**

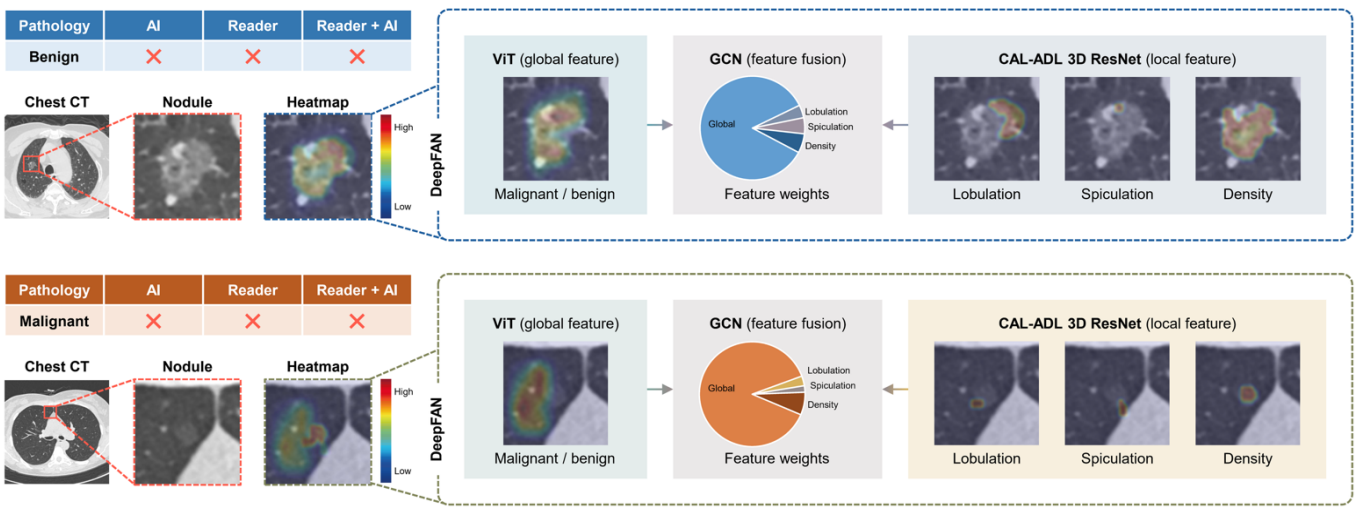